\documentclass[ba]{imsart}
\pubyear{0000}
\volume{00}
\issue{0}
\doi{0000}
\firstpage{1}
\lastpage{1}
\usepackage{booktabs}
\usepackage{float}
\usepackage{multirow}
\usepackage{booktabs}
\usepackage{adjustbox}
\usepackage{float}
\usepackage{amsthm}
\usepackage{amsmath}
\usepackage{natbib}
\usepackage[colorlinks,citecolor=blue,urlcolor=blue,filecolor=blue,backref=page]{hyperref}
\usepackage{graphicx}
\usepackage{amsfonts}
\usepackage{amssymb}
\usepackage{algpseudocode}
\usepackage{mathtools, algorithm, algorithmicx}
\usepackage{listings}
\usepackage{color}
\usepackage{multirow}
\usepackage{caption}
\usepackage{subcaption}
\usepackage{graphicx}
\graphicspath{ {./images/} }
\usepackage{amssymb}
\usepackage[utf8]{inputenc}
\usepackage{lipsum}
\usepackage{mathtools}
\usepackage{cuted}
\usepackage{comment}
\startlocaldefs
\numberwithin{equation}{section}
\theoremstyle{plain}
\newtheorem{thm}{Theorem}[section]

\newtheorem{rem}{Remark}
\newtheorem{prop}[thm]{Proposition}

\endlocaldefs

\begin{document}

\begin{frontmatter}
\title{Efficient Model Compression for Bayesian Neural Networks}
\runtitle{Efficient Model Compression for BNN}

\begin{aug}
\author{\fnms{Diptarka} \snm{Saha}\ead[label=e1]{saha12@illinois.edu}},
\author{\fnms{Zihe} \snm{Liu}\ead[label=e1]{saha12@illinois.edu}},
\author{\fnms{Feng} \snm{Liang}\ead[label=e2]{liangf@illinois.edu}}


\end{aug}

\begin{abstract}

Model Compression has drawn much attention within the deep learning community recently. Compressing a dense neural network offers many advantages including lower computation cost, deployability to devices of limited storage and memories, and resistance to adversarial attacks. This may be achieved via weight pruning or fully discarding certain input features. Here we demonstrate a novel strategy to emulate principles of Bayesian model selection in a deep learning setup. Given a fully connected Bayesian neural network with spike-and-slab priors trained via a variational algorithm, we obtain the posterior inclusion probability for every node that typically gets lost. We employ these probabilities for pruning and feature selection on a host of simulated and real-world benchmark data and find evidence of better generalizability of the pruned model in all our experiments. 
\end{abstract}


\end{frontmatter}
\section{Introduction}

 Neural network based pattern learning techniques have influenced modern technology like very few things before it has. One of the biggest advantages that neural network models enjoy over traditional statistical learning methods is their ability to approximate, to any degree of precision, any functional form that may be present in the inherent data generation method \citep{HORNIK1991251}. This flexibility has allowed large-scale democratization of such methods \citep{lecun, Goodfellow-et-al-2016}. 
\par
The flexibility or complexity of a neural network model is controlled by its architecture (i.e., the depth and/or width of each layer). Although a `large' or over-parametrized network can approximate any data-generating function to arbitrary precision, there are several drawbacks of an over-parametrized model. First of all, over-parameterised models tend to overfit the data and degrade generalization to unseen examples \citep{bartoldson2020generalization}; for example, experiments in \cite{zhang2021understanding} have shown that it is easy to train neural networks to fit random labels and such models, of course, have generalization errors no better than random guessing. Although research has shown that over-parameterized neural networks can achieve good generalization performance due to implicit regularization imposed by training algorithms such as early stopping or stochastic gradient descent, the interplay among model complexity, regularization, and the generalization error of a large neural network model has still not been fully understood \citep{belkin2019reconciling,zhang2021understanding}. Further, 
a large model comes at the cost of additional memory and computation effort during model training and inference \citep{hoefler2021sparsity}. Sparse models, on the other hand, are easier to store with computational savings and are more robust against adversarial attacks \citep{10.1145/3386263.3407651}. \par

Techniques used to reduce the complexity of a neural network while retaining much or all of its predictive power are, collectively, termed as  \emph{model compression}. Systematic reviews, such as \cite{cheng2020survey}, categorize recent developments in model compression for deep neural networks into four broad categories: parameter pruning, low-rank factorization, transferred convolutional filters,  and knowledge distillation. In this paper, we focus on \emph{parameter pruning}, this technique entails compressing an existing, typically large and accurate, trained network by systematically removing parameters from it while keeping the accuracy unharmed. Pruning has quite a rich history \citep{PhysRevA.39.6600, 10.5555/2969735.2969748, 369200}; however, it has seen an explosion of interest in the past decade. Within the framework of pruning, it is common to assign a score for relevance to each parameter; based on the scores, we then remove individual parameters (\textit{unstructured pruning or weight pruning}). 
\citet{blalock2020state} offers an excellent review of the state of pruning for neural networks.

We will employ a Bayesian approach to neural network pruning -- it not only provides valid uncertainty estimates but also allows for the concurrent estimation of regression coefficients and complexity parameters. As  \citet{rockova2012hierarchical} puts it, Bayesian regularization falls within the versatile framework of Bayesian hierarchical models.  In particular, we formulate the problem of parameter pruning as a special case of Bayesian variable selection \citep{Ohara, clyde2004}. We consider structured pruning, removing a weight/neuron (\textit{weight/node pruning}) or an entire feature (\textit{feature reduction}/\textit{variable pruning}).
In the Bayesian framework, the problem of parameter pruning may be viewed as a special case of variable selection.
There have been many notable works on the theory and application of feature selections in a BNN recently \citep{LiangFS, DinhFS, DeepFS, jzliu21g}. In the broader Bayesian literature, the gold standard for Bayesian variable selection revolves around the use of spike-and-slab priors \citep{Mitchell1988, SSVS, Smith, clyde, George1997, Ishwaran_2005}.
This is convenient as it can seamlessly integrate model averaging and variable selection within a single estimation process.
In this strategy, we rely on an auxiliary indicator variable $Z$, which denotes if the slab or spike part of the prior is dominant to infer whether the feature is included in the model or not, respectively. Thus, direct inference on the distribution of $Z$ is imperative. 
\par 
Bayesian methods have been used in neural networks \citep{Neal, Mackay}. Bayesian neural networks (BNNs)  present distinct advantages over regular neural networks by offering a natural mean for uncertainty quantification and model interpretation. In practice, however, BNNs suffer from high computational costs as exact Bayesian inference on the weights of a neural network is intractable and decent approximations are resource intensive.
These computational challenges further exacerbate when dealing with discrete quantities of interest, such as $Z$ when using the spike-and-slab prior. The main contribution of our work is an efficient algorithm that learns not just a Bayesian neural network's weights but also its topology directly by approximating the posterior of the inclusion variables. 

Before diving into the main content of our paper, we first briefly outline the distinctions between our work and that of two relevant papers.
\begin{itemize}
\item 
The strategy of using inclusion probabilities for regularisation purposes has been the cornerstone of the \textit{dropout} technique \citep{srivastava14a}. However, there are two key differences between our method and these works: in dropout, (i) the sparsity is layerwise and doesn't apply to weights individually, and (ii) more importantly, the sparsity is fixed beforehand and not learned from the data itself. Thus our work may be viewed as an \textit{adaptive} version of dropout 
In fact, \cite{gal2016uncertainty_1} has argued that using a spike-and-slab prior may be seen as a particular form of dropout training; he further mentioned that typical dropout training is non-adaptive and hence a grid search may be required. Here we can reach the same goal of learning the topology of the network directly. 
\item Another work that we feel is closely related to ours is the variational approximation technique \textit{Bayes by Backprop} from \cite{blundell2015weight}. They do accommodate spike-and-slab priors on the weights but avoid the computational challenges on the posterior by using mean field Gaussian variational distribution -- this camouflages the potential sparsity in the data. We remove this restriction in this work to learn and leverage this sparsity. 
\end{itemize}



The remainder of this paper is arranged as follows. Section \ref{sec:bg} discusses the background: we lay down the setup in Section \ref{sec:bg:setup}, discuss the variational approach to BNN training in Section \ref{sec:bg:VI}, provide a brief review of pruning techniques in BNNs in Section \ref{sec:bg:priorwork},  and finally in Section \ref{sec:bg:motivation}, collate the gap in the literature and discuss how we intend to address it. Section \ref{sec:meth} provides the main contribution that we combine into one algorithm for training a sparse BNN and a feature relevance detection strategy defined. Finally, Section \ref{sec:exp} discusses all the experiments. We conclude the paper with a discussion and future prospects in Section \ref{sec:conclusion}.

\section{Background} \label{sec:bg}
\subsection{Setup} \label{sec:bg:setup}

Given a dataset $\mathcal{D}$ consisting $n$ independent observations of  $\mathbf{x} = (x_1, \dots, x_p)$ and target $y$, we are interested in learning their relationship so that given a new data-point $x^*$ we are able to provide an accurate prediction $y^*$. For a neural network with $L$ hidden layers, the estimated model has the following form: 
\begin{equation} \label{eq:f-hat:NN}
\hat{y} = f_\mathbf{W} (\mathbf{x}) =  \sigma_L \cdots \sigma_2 (\sigma_1 (\mathbf{x}^T w_1 + b_1) ^T w_2 + b_2 )\cdots ). 
\end{equation}
where 
$\{\sigma_j\}_1^L$ are (typically non-linear) activation functions specified by the user. For example, the ReLU (rectified linear unit) activation function is defined as: $\sigma(z) = \max(z, 0).$ The complexity of the model (\ref{eq:f-hat:NN}) can be further increased by increasing the number of layers and/or nodes in each layer. The goal then is to train the network such that some loss is minimized; for example,  for regression tasks, we might be interested to minimise the euclidean loss between actual and predicted values, while for classification tasks we may be interested in minimizing the cross-entropy loss.
\par 
Although methods like these have been successful in complicated pattern recognition tasks, they have remained statistical black boxes lacking critical information needed for inference such as uncertainties associated with the predictions. Bayesian methods offer a natural way to reason about uncertainty in predictions and can provide insight into how these decisions are made. To do this, a distribution is placed over the network
parameters and the resulting network is then termed a Bayesian Neural Network (BNN). 

Formally, consider a BNN with $L$ layers and $n_l$  nodes in each layer $l \in \{1, ..., L\}$. Let $\textbf{W} = \{w_j, b_j\}_1^L$ denote the corresponding parameters where $w_j$ denotes the $n_{l-1} \times n_l$ weight matrix, $b_j$ denotes the $n_{l} \times 1$ bias vector, and $n_0 = p$, the input feature dimension. Let $\pi(\cdot)$ denote the prior on $\textbf{W}$. The objective, then, will be to compute the resulting posterior distribution:
\begin{equation} \label{eq:post(w,z)}
\pi(\mathbf{W}|\mathcal{D}) =  \frac{\pi(\mathbf{W})p(\mathcal{D}|\mathbf{W})}{p(\mathcal{D})} .
\end{equation}
 \citet{Neal} popularised such a setup with Gaussian priors for his 
 networks. In the past decade, authors have employed several priors that help their use-case including  mixed Gaussian  \citep{blundell2015weight}, Horseshoe  \citep{louizos2017bayesian}, etc. 

\par
Leveraging this distribution, we can predict output for a new input point $x^*$ by integrating: 
$$p(y^* |x^*, \mathcal{D}) = \int p(y^* |x^* , w) p(w|\mathcal{D})dw. $$
Note, since the output is an entire distribution and not just one value, we can easily proceed to investigate any statistic we are interested in, including the mean and uncertainty estimates among others. This advantage has promoted BNNs to prominence as the explainability of artificial intelligence has emerged to be of paramount interest, especially security-critical applications \citep{bykov2021explaining}.

\subsection{Approximate Posterior via Variational Inference}
\label{sec:bg:VI}

Baring certain ideal cases of conjugate prior, the posterior \eqref{eq:post(w,z)} is not analytically tractable. Thus the problem of numerically approximating the posterior is imperative. Some early notable works include \cite{Neal}'s  hybrid Monte Carlo or Hamiltonian Monte Carlo; \cite{Mackay}'s Laplace method; and \cite{hinton1993}'s MDL-based approach under a variational inference interpretation, which was further generalized by \cite{barber1998ensemble}. Please see \cite{goan2020bayesian} for a thorough review.

In this work, we focus on variational inference (VI). VI frames marginalization required during Bayesian inference as an optimization problem. This is achieved by assuming the form of the posterior distribution and performing optimization to find the assumed density closest to the true posterior. This assumption simplifies computation and provides some level of tractability.

The assumed variational distribution $q_\theta (\cdot)$ is a suitable density over the set of parameters $\textbf{W}$, which is restricted to a certain family of distributions parameterized by $\theta$. The parameters for this variational distribution are then adjusted to reduce the dissimilarity between the variational distribution and the true posterior. The dissimilarity is often measured by their KL divergence:
\begin{align*}
    \text{KL} (q_\theta (\textbf{W}) ||\pi(\textbf{W}|\mathcal{D}) ) &= \int 
q_\theta (\textbf{W}) \log \frac{q_\theta (\textbf{W})}{\pi(\textbf{W}|\mathcal{D}) } d\mathbf{W} \\
& = \int 
 q_\theta (\textbf{W}) \log \frac{q_\theta (\textbf{W})}{\pi(\textbf{W}) p(\mathcal{D} | \textbf{W}) / p(\mathcal{D})} d\mathbf{W} \\
&= \text{KL} (q_\theta ( \textbf{W}) || \pi(\textbf{W}) ) - \mathbb{E}_{q_\theta (\textbf{W})} \log p(\mathcal{D}| \textbf{W}) + \log p (\mathcal{D}).
\end{align*}
Note that the last term is independent of $\theta$.  Thus, finding the optimal variation parameter is equivalent to maximizing the following objective function:
\begin{equation} \label{eq:VI}
   - \text{KL} (q_\theta (\textbf{W}) || \pi(\textbf{W}) ) + \mathbb{E}_{q_\theta (\textbf{W})} \log p(\mathcal{D}| \textbf{W}).
\end{equation} 
This objective function, often termed  the Evidence Lower BOund (ELBO), contains two antagonistic terms: a prior-dependent part and a data-dependent part. Thus finding an optimal variational posterior $q_\theta (\cdot )$ can be thought of as finding the balance between the complexity of the data $\mathcal{D}$ and simplicity of the prior $ \pi(\cdot)$.

Gradient descent is often used to optimize \eqref{eq:VI}.  However, computing the gradient of the second term is often analytically impossible and computationally challenging. To alleviate the issue, \cite{REINFORCE} uses the so-called \textit{log-derivative} trick, which
leverages the differential rule, $\nabla_\theta p_\theta(x) = p_\theta(x) \nabla_\theta \log p_\theta(x)$, for any density $p$ parameterised by $\theta$. 
This trick is easy to implement and often quite useful, especially if the density in question belongs to the exponential family. However, the Monte Carlo estimates of the gradients computed using such methods suffer from high variance, which necessitates higher resources for computation. Alternatively, \cite{blundell2015weight} has outlined a general pragmatic approach to optimize the variational parameters,
which relies heavily on the \textit{reparametrization trick} \citep{Kingma2014AutoEncodingVB}. This method yields low variance but is restricted only to the continuous family of distributions. 
More details on this technique will be discussed in Section \ref{sec:meth}.



\subsection{Prior Work on Bayesian Pruning}
\label{sec:bg:priorwork}
The earliest idea of pruning a neural network was introduced by \cite{Cun90optimalbrain}. \cite{Han2016DeepCC} have applied the same idea to modern architectures. Simple heuristic methods removing weights of small magnitude have also been employed with great compression ratio but without much theoretical justification \citep{Strom1997SparseCA, CollinsK14}.  The issue has also been targeted from a Bayesian perspective in works such as \cite{Neal}, \cite{Mackay}, etc. While theoretically sound, these results lack scalability due to the difficulty of obtaining or estimating the posterior. 

Perhaps, in the past decade, the most popular and empirically effective method for regularising a neural network has been the \textit{Dropout} method \citep{srivastava14a}. In this method, we add multiplicative noise to the input of each layer of the neural network which makes the weights less likely to overfit \citep{Kingma2015VariationalDA}. In their seminal work, \citet{gal2016dropout} added Bayesian justification for this approach by establishing it as an approximation to a Gaussian Process. Despite its success, initial versions of dropout had some major drawbacks: working with binary errors is expensive; but more importantly, the sparsity parameters (dropout rates) are fixed and not learned from the data. Further to avoid potential complexity the rates were also shared across layers, which masks the sparsity of individual weights and nodes. 

In \emph{Gaussian Dropout}, \citet{pmlr-v28-wang13a} alleviated the first issue by successfully showing that one can reap all the benefits of the dropout by using an appropriately selected Gaussian noise instead of a binary one. In \textit{Variational Dropout},  \citet{Kingma2015VariationalDA} re-imagined the Gaussian Dropout as a variational approximation procedure with 
Jeffrey's prior $\pi(w) \propto \frac{1}{w}$ on the weights and the variational posterior $q(w) \sim N(\theta, \alpha \theta^2)$. Manipulating the variational objective function will allow us to learn the individual dropout rates for each layer, node, or weight -- facilitating a sparse solution. The idea has been further modified and verified empirically in \cite{molchanov2017variational}. Extensions of similar ideas have been demonstrated in \cite{louizos2017bayesian}, 
where the authors used the same variational posterior as in \citet{Kingma2015VariationalDA}, but a scaled mixture of Gaussian prior to achieve Bayesian compression.

\subsection{ Motivation}
\label{sec:bg:motivation}

The spike-and-slab prior remains the gold standard for a Bayesian relevance determination of a parameter. However, approaches for pruning a BNN have avoided inferring the inclusion probability of any weight due to the associated complexity in estimating the gradient. The difficulties stem from the fact that the reparameterization trick becomes inapplicable if the quantities of interest have discrete distribution simply due to the fact that the objective function thereof ceases to be differentiable and hence estimation of the gradient is impossible. This presents a conundrum -- in order to induce sparsity we must infer about \textit{inclusion variables} that will take binary (discrete) values. Algorithms recruiting the reparameterization trick, such as \textit{Bayes by Backprop} from \cite{blundell2015weight}, will fail to do so. 

In their original work,  \cite{blundell2015weight} 
has only used mean field Gaussian posterior but  demonstrated a possible strategy for pruning by looking at the signal-to-noise ratio of each weight. While the empirical utility of such rule-of-thumb techniques is well-established, they do not provide a principled way to estimate the posterior inclusion probability of any weight. In \cite{molchanov2017variational} and extensions thereof,  authors explicitly use a continuous mixture of Gaussian priors continuous  instead of discrete spike-and-slab prior to reduce computational expense. Investigations into the application of the reparameterization trick for discrete parameters are still nascent at this time. Some approaches include the use of techniques such as marginalization \citep{tokui2016reparameterization}, Gaussian approximation \citep{shayer2018learning}, and the Straight-through estimator \citep{courbariaux2016binarized, li2016ternary, rastegari2016xnornet}. 

In this work, we employ the spike-and-slab prior to the weights. For the posterior, unlike previous works, we assume multiplicative mean field Bernoulli (for inclusion probabilities) and Gaussian (for weight sizes) distributions. The aforementioned issues are bypassed by leveraging nice relationships among the concerned parameters arising in this particular case. By estimating not just the Gaussian but also the Bernoulli parameters, we can learn the optimal model topology with probabilistic guarantees. We leverage these probabilities to generate strategies that facilitate weight pruning and feature pruning. 

\section{Method}
\label{sec:meth}

\subsection{A New VI Objective Function}
\label{sec:meth1}

The architecture of any neural network is heuristic at best;  thus redundancies are likely to prevail. In Bayesian literature, a common weapon to identify and eliminate such redundancies is the sparsity-inducing spike-and-slab prior. We will assume the same here for all weights and biases with a unique sparsity-inducing variable $Z_i$ for each parameter. The spike-and-slab prior can be written as follows:
\begin{align}
  \pi (\mathbf{W}, \mathbf{Z})  = \prod_{\mathbf{W}, \mathbf{Z}} \Big [ \pi \cdot \mathcal{N} (w_i; 0, \tau^2_1 ) \Big ]^{Z_i} \cdot \Big [(1-\pi) \cdot \mathcal{N} (w_i; 0, \tau^2_0 ) \Big ]^{(1- Z_i)}   \label{eq:prior}
\end{align}
where $\tau^2_0 < \tau_1^2$. Note that if we integrate over the binary latent variables $\mathbf{Z}$, we retrieve the marginal prior distribution on weights used in \cite{blundell2015weight}: 
$$ \pi (\mathbf{W})  = \prod_{\mathbf{W}} \Big [ \pi \cdot \mathcal{N} (w_i; 0, \tau^2_1 ) +  (1-\pi) \cdot \mathcal{N} (w_i; 0, \tau^2_0 ) \Big ]. $$ 
The objective, then, will be to compute the resultant posterior distribution
\begin{equation} 
\pi(\mathbf{W, Z}|\mathcal{D}) \propto   \pi(\mathbf{W, Z})p(\mathcal{D}|\mathbf{W}).
\end{equation}

As mentioned in Section \ref{sec:bg:motivation}, the existing scalable methods exclude  discrete variables such as $\mathbf{Z}.$ We intend to bridge this gap by assuming a separate variational distribution on the latent binary inclusion variable along with the network weights:
\begin{equation} \label{eq:q:w:Z}
q_{\mathbf{p}}(\mathbf{Z}) = \prod_{\mathbf{Z}} Bern(Z_i; p_i), \quad q_{\theta}(\mathbf{W}) =  \prod_{\mathbf{W}} \mathcal{N}(w_i; m_i, \sigma_i^2),
\end{equation}
where $\theta = \{m_i, \sigma_i^2\}_{i=1}^M$ and $\mathbf{p} = \{p_i\}_{i=1}^M$ denote all the variational parameters associated with $\mathbf{W}$ and $\mathbf{Z}$, respectively. We may refer to $\theta$ as \textit{`weight parameters'} and $\mathbf{p}$ as \textit{`sparsity parameters'}. 

Our variational objective function then becomes
\begin{align}
   \mathcal{J}(\theta, \mathbf{p})  &= - \mathbb{E}_{q_{\theta}(\mathbf{W})} \mathbb{E}_{q_{\mathbf{p}}(\mathbf{Z})}  \log \frac{p(\mathcal{D} | \mathbf{W}) \cdot \pi(\mathbf{W}, \mathbf{Z})}{q_{\theta} (\mathbf{W}) \cdot q_{\mathbf{p}}(\mathbf{Z})} \label{eq:masterOF}.
\end{align}
This objective function explicitly optimizes the distribution of the inclusion variable while making it customizable for each weight individually and providing the optimal sparsity for each weight. This constitutes a stark difference from \textit{Dropout} and \textit{Bayes by Backprop}, which is the core of our work.  
\par 
To compute the gradient of  \eqref{eq:masterOF}, we use 
the reparametrization trick \citep{Kingma2014AutoEncodingVB}, which is, in essence, very similar to pivoting technique \citep{casella2002statistical} used in classical statistics. This trick hinges on the fact that a random variable may be represented as a combination of a deterministic and a differentiable expression.
For example,  with $\theta = \{\mu, \sigma^2 \}$, if 
$\omega \sim N (\mu, \sigma^2 )$ and $\epsilon \sim N (0,1) $, then 
    \begin{align}
           \omega = g (\theta, \epsilon) = \mu + \sigma \cdot \epsilon \label{eq:repam}.
    \end{align}
This allows us to make the expectations in \eqref{eq:masterOF} free of parameters $\theta$ by rewriting them in terms of this simpler, differentiable random variable $\epsilon$. 

Complications arise in the optimization process with the discrete random variables $Z_i$, as they cannot be reparametrised as a differentiable function of a continuous random variable. Thus, the objective function ceases to be differentiable. However, the following result notes that the dependence on $Z_i$'s variational parameter $p_i$ is isolated only to a non-stochastic penalty term. Thus, we can bypass the computational challenges and will still be able to use the reparametrization trick. 

\begin{prop} \label{lemma:J:two-parts} The objective function \eqref{eq:masterOF} can be alternatively expressed as
\begin{align}
      \mathcal{J}(\theta, \mathbf{p})  &=  -\mathbb{E}_{q_{\theta}(\mathbf{W})} \log p(\mathcal{D} | \mathbf{W}) + \sum_{i} \mathcal{R}(\theta_i, p_i ) \label{eq:solution}
\end{align}
up to a constant, where
\begin{align}
\mathcal{R}(\theta_i, p_i) =  p_i \cdot\Big[\frac{m_i^2 + \sigma_i^2}{2\tau_1^2} + \log \frac{\tau_1p_i}{\sigma_i\pi}\Big ] + (1-p_i)\cdot \Big[\frac{m_i^2 + \sigma_i^2}{2\tau_0^2} + \log \frac{\tau_0(1 - p_i)}{\sigma_i(1-\pi)} \Big]. \label{eq:R:theta_i}
\end{align}
\end{prop}

Note that $p_i$'s, the variational parameters associated with the discrete latent variables $Z_i$'s, appear only in the second term of the objective function \eqref{eq:solution}. As demonstrated in the following result, for any given set of weight parameter values $\theta_i = (m_i, \sigma_i^2)$, the optimal value of the sparsity parameter $p_i$ can be expressed as a closed-form function of $\theta_i$. 


\begin{prop} \label{lemma:p:function}
For any set of values of the weight parameters $({m}_i, {\sigma}_i^2 )$, the optimal value of the sparsity parameter $p^*_i$ satisfies the following equality: 
\begin{align}
\label{eq:p_i}
p^*_i(\theta_i) = \frac{1}{1 + \exp \{ A_i - B_i \}},
\end{align}
where
\begin{align*}
    A_i = \frac{m_i^2 + \sigma_i^2}{2\tau_1^2} + \log \frac{\tau_1}{\pi},  \quad 
    B_i = \frac{m_i^2 + \sigma_i^2}{2\tau_0^2} + \log \frac{\tau_0}{1-\pi}. 
\end{align*}


\end{prop}

The proofs of these results may be found in the supplementary material. 

\subsection{Algorithm to Optimise $J$}\label{sec:meth2}


The above two results, Proposition \ref{lemma:J:two-parts} and Proposition \ref{lemma:p:function}, suggest a coordinate descent algorithm for the optimization of\eqref{eq:masterOF}: at every iteration, given the sparsity parameters $\mathbf{p}$, update $\theta$ via gradient descent; then given the weight parameters $\theta$, update $\mathbf{p}$ via the closed-form function \eqref{eq:p_i}.

Optimization over $\theta$ can be done using the reparametrization trick. First,  let $\mathcal{R}(\theta_i)$ denote $\mathcal{R}(\theta_i, p_i)$, which is a function of $\theta_i$ only when  $\textbf{p}$ is given. Then write the objective function \eqref{eq:solution} as 
\begin{equation}
\mathcal{J} (\theta)= \mathbb{E}_{q_\theta (\mathbf{W})} \big [ -\log p(\mathcal{D} | \mathbf{W}) + \sum_i \mathcal{R}(\theta_i) \big ].
\label{eq:J(theta)}
\end{equation}
Define $f(\theta, \mathbf{W}) = -\log p(\mathcal{D} | \mathbf{W}) + \sum_i \mathcal{R}(\theta_i).$
Following \eqref{eq:repam}, we can write  for this differentiable function $f$,
\begin{align*}
    \mathbb{E}_{q_\theta (\mathbf{W})} \big [f(\theta, \mathbf{W}) \big ] = \mathbb{E}_{\phi(\boldsymbol{\epsilon})} \big [f(\theta, g (\theta, \boldsymbol{\epsilon})) \big ] = \int f(\theta, g (\theta, \boldsymbol{\epsilon})) \phi(\boldsymbol{\epsilon}) d \boldsymbol{\epsilon}, 
\end{align*}
where $\mathbf{W} = g(\theta, \boldsymbol{\epsilon})$ is the multivariate version of \eqref{eq:repam} and $\phi(\cdot)$ denotes the standard multivariate Gaussian density function. 
Since $\boldsymbol{\epsilon}$ is free of $\theta$, the integral and the gradient operators can be interchanged:
\begin{align}
\nabla_\theta  \mathbb{E}_{q_\theta (\mathbf{W})} \big [f(\theta, \mathbf{W}) \big ] = \nabla_\theta \mathbb{E}_{\phi(\boldsymbol{\epsilon})} \big [f(\theta, g (\theta, \boldsymbol{\epsilon})) \big ] = \mathbb{E}_{\phi(\epsilon)} \big [ \nabla_\theta f(\theta, g (\theta, \boldsymbol{\epsilon})) \big ]. 
 \label{eq:repam3}
\end{align}
This gradient can be numerically approximated. We can sample $L$ many i.i.d. samples of $\boldsymbol{\epsilon}$ from standard multivariate Gaussian distribution and find the Monte Carlo estimate:
\begin{align}
    \mathbb{E}_{\phi(\boldsymbol{\epsilon})} [ \nabla_\theta f (\theta, g(\theta, \boldsymbol{\epsilon}))] &\approx \frac{1}{L} \sum_{l = 1}^L \nabla_\theta f (\theta, g(\theta, \boldsymbol{\epsilon}^{(l)} )) \label{eq:repam_MC}.
\end{align}
Following \eqref{eq:repam3} and \eqref{eq:repam_MC}, the gradient for the objective function may be calculated and approximated as 
\begin{align}
    \frac{\partial}{\partial \theta}\mathcal{J} (\theta) =  \mathbb{E}_{\phi(\boldsymbol{\epsilon}) }    \Bigg[ \frac{\partial}{\partial \theta} f \left(\theta,\mathbf{W}\right) \Bigg]
    \approx \frac{1}{L} \sum_{l = 1}^L \Bigg[ \frac{\partial f}{\partial \mathbf{W}} \frac{\partial \mathbf{W}}{\partial \theta}  +\frac{\partial f}{\partial \theta}  \Bigg] \label{eq:blun_MC}.
\end{align}

 This optimization process is summarised in the following automated algorithm. Here,  the variational parameters are $\theta = \{ m,  \rho \}$ and $\sigma = \log (1 + e^\rho)$; this is to make sure we have positive variance. The output of the algorithm is the mean, and variance of the final Gaussian distribution for each weight, and its probability of coming from the slab component.

\begin{algorithm}[H] 
\caption{Sparse BNN Approximation}
\begin{algorithmic}[1]\label{algo:1}
\Procedure{sBNN}{$\theta, \mathcal{D}, \alpha$}
\Repeat
 \State \textbf{Sample: $\epsilon \sim N (\mathbf{0}, \mathbf{I})$ }
     \State Let $\theta = (\mathbf{m}, \boldsymbol{\rho})$
     \State Set $\mathbf{W} = \mathbf{m} + \log (1 + \exp (\boldsymbol{\rho})) \odot \epsilon $
    \State \textbf{Compute} $f(\mathbf{W},\theta, p ) = -\log p(\mathcal{D} | \mathbf{W}) + \mathcal{R}(\theta, p) $
    \State \textbf{Compute} the gradients using reparametrization technique
    \begin{align*}
        \nabla_m &= \frac{\partial f}{\partial \mathbf{W}} + \frac{\partial f}{\partial m}
        \\
        \nabla_\rho &= \frac{\partial f}{\partial \mathbf{W}} \frac{\epsilon}{ 1 + \exp(-\rho)} + \frac{\partial f}{\partial \rho} 
    \end{align*}
    \State \textbf{Update the weight parameters} using gradient descent
    \begin{align*}
        m &\longleftarrow m - \alpha \nabla_m
        \\
        \rho &\longleftarrow \rho - \alpha \nabla_\rho
    \end{align*}
    \State \textbf{Update the sparsity parameter} using \eqref{eq:p_i}
    \begin{align*}
        \sigma  &\longleftarrow \log(1 + \exp (\rho)) \\
        A  &\longleftarrow  \frac{{m}+ {\sigma}^2}{2\tau_1^2} + \log \frac{\tau_1}{\pi}\\  
        B  &\longleftarrow \frac{{m}^2 + {\sigma}^2}{2\tau_0^2} + \log \frac{\tau_0}{1-\pi}\\
        p &\longleftarrow  \frac{1}{1 + \exp \{ A - B \}}
    \end{align*}
\Until{Convergence}
\EndProcedure
\State \textbf{Output}: $\{m_i, \sigma_i^2, p_i\}$ for each weight $w \in \textbf{W}$
  \end{algorithmic}
\end{algorithm}

\subsection{Comparing Objective Functions} \label{sec:compare:obj}


Incorporating the sparsity parameter indirectly into the objective function creates some compelling parallel with the \emph{Bayes by Backprop} method from \cite{blundell2015weight}. In this section, we compare the objective function we are solving, denoted by $\mathcal{J}$ defined in \eqref{eq:J(theta)}, to that from \cite{blundell2015weight}, denoted by $\mathcal{J}_B$:
\begin{eqnarray*}
     \mathcal{J}(\theta) & =  & \mathbb{E}_{q_{\theta}(\mathbf{W})} \Big [ - \log p(\mathcal{D} | \mathbf{W})  + \sum_i \mathcal{R}(\theta_i)\Big ]\\
    \mathcal{J}_B(\tilde{\theta}) & = & \mathbb{E}_{q_{\tilde{\theta}} (\mathbf{W})} \Big [ - \log p(\mathcal{D} | \mathbf{W}) - \log \frac{\pi(\mathbf{W})}{q_{\tilde{\theta}}(\mathbf{W})} \Big],
\end{eqnarray*}
where two variational distributions $q_{\tilde{\theta}}$ and $q_{\theta}$ on $\mathbf{W}$ take the same form as  defined in \eqref{eq:q:w:Z}.

In particular, we compare their gradients. Let's drop the index without loss of generality and  only compare for one weight $W$.  Since the first term is the same in both these functions, we compare the second term, which is 
\begin{align*}
    f(W , m) = - \log \pi (W) + \log q (W| m)
\end{align*}
in  $J_B$  and $\mathcal{R}(\theta)$ in our objective function $J$. See the Supplementary Material for detailed derivation.

First, let us compare the gradient with respect to the mean parameter $m$. 
An MCMC Estimate of $\nabla_m f$ in \cite{blundell2015weight} is given by
\begin{align}
    \hat{\nabla}_m f = \frac{\hat{W}}{\tau_1^2} \frac{\pi_1(\hat{W})}{\pi(\hat{W})} + \frac{\hat{W}}{\tau_0^2} \frac{\pi_0(\hat{W})}{\pi(\hat{W})} 
    \label{eq:gradient:J_B:m}
\end{align}
where $\hat{W} = m + \sigma \hat{\epsilon}$ is a sample from  $N (m, \sigma^2)$.  On the other hand, the gradient of $\mathcal{R}(\theta)$ in our approach is 
\begin{align}
  \nabla_m \mathcal{R}(\theta) = \frac{m}{\tau_1^2} p + \frac{m}{\tau_0^2} (1-p) = \frac{m}{\tau_1^2} \frac{\pi_1(m)}{\pi(m)} + \frac{m}{\tau_0^2} \frac{\pi_0(m)}{\pi(m)}.
  \label{eq:grad_vs_1}
\end{align}
These two gradients are directly comparable,  notice $\dfrac{\pi_1(W)}{\pi(W)} = \mathbb{E} (Z | W)$ and $p = \mathbb{E}_{q(Z)} Z$. Since we have explicitly obtained distribution for $Z$, we can use its expectation directly. However, in \cite{blundell2015weight}'s case they have to use its conditional expectation (and estimates thereof) given a weight value since $Z$ doesn't have a direct distribution. Similarly, our objective function substitutes the variational mean $m \mathbb{E}_{q(W)} W$  of $W$, instead of the whole random variable.  

For the variance parameter $\sigma^2$ similar comparisons exist. In our case, the gradient is 
\begin{align}
    \nabla_{\sigma^2} \mathcal{R}(\theta) =  \frac{1}{2} \left( \frac{1}{\tau_1^2} p + \frac{1}{\tau_0^2} (1-p)  - \frac{1}{\sigma^2} \right),  \label{eq:grad_vs_2}
\end{align}
while an MCMC estimate of the gradient in \cite{blundell2015weight} is given by
\begin{equation}
    \hat{\nabla}_{\sigma^2} f = \frac{1}{2} \left(\frac{m}{\sigma}  \hat{\epsilon} + \hat{\epsilon}^2 \right )   \Bigg(\frac{1}{\tau_1^2} \frac{\pi_1(\hat{W})}{\pi(\hat{W})} + \frac{1}{\tau_0^2} \frac{\pi_0(\hat{W})}{\pi(\hat{W})} - \frac{1}{\sigma^2} \Bigg) +\frac{1}{2\sigma^2} \left(  \hat{\epsilon}^2 -1  + \frac{m}{\sigma} \hat{\epsilon} \right)  \label{eq:grad_bl_2}
\end{equation}
From the previous discussion regarding the mean parameter, it is clear that the middle term in \eqref{eq:grad_bl_2} is analogous to the whole gradient in \eqref{eq:grad_vs_2}. The other two terms are random, and in \cite{blundell2015weight}'s algorithm, these are estimated via MCMC. However, these terms also have the following nice properties that help us see the comparison through:
$$  \mathbb{E} \left( \frac{m}{\sigma}  \hat{\epsilon} + \hat{\epsilon}^2 \right)  = 1 , \quad 
\mathbb{E} \left(  (\hat{\epsilon}^2 -1 ) + \frac{m}{\sigma} \hat{\epsilon}\right) = 0.
$$

In summary, we observe that the gradients have a very similar form in both, while certain quantities of interests have different but analogous estimates. Most importantly, for the distribution of $Z$, we  use the exact expectation given by $p_i$ from \eqref{eq:masterOF}, while \textit{Bayes by Backprop}  uses its conditional expectation given a random sample of the weight. This highlights the core difference, and showcases how even though the ideas are along similar lines, we can leverage explicitly the variational distribution of the sparsity parameter that we have employed.

\subsection{Remarks on the Final Solution}
Finally, let us discuss some repercussions of using this sparsity parameter learned through our algorithm. To do so, first, we observe, 
the final solution as described in  (\ref{eq:p_i}) can also be written as 
\begin{align}
    2 \bigg(\text{logit}(p_i) - \text{logit} (\pi_i) \bigg) = \left(m_i^2 + \sigma_i^2\right)\left(\frac{1}{\tau_0^2}  - \frac{1}{\tau_1^2}  \right) - \log \frac{\tau_1^2}{\tau_0^2}. \label{eq:p_i2}
\end{align}

Using this form, we can now remark on some interesting characteristics of our estimate of the inclusion probability. 

\begin{rem}
 For any given prior variances $\tau_1^2$ and $\tau_0^2$, 
 $p_i$ is a monotonically increasing function of $m_i^2 + \sigma_i^2$.
\end{rem}
 This is a desirable property. A larger magnitude of the second moment indicates that the slab component is likely to be dominant, and thus $p$ should be higher. We can then prune off weights with low-second moments. Note, this is somewhat different from pruning based on signal-to-noise ratio $m_i/\sigma_i$ as originally prescribed in \citet{blundell2015weight}. This property also indicates: $p \to 1$ when  $m_i^2 + \sigma_i^2 \to \infty.$

\begin{rem}
 As $\dfrac{\tau_1^2}{\tau_0^2}\to 1$, we have $\dfrac{p}{\pi} \to 1 $.
\end{rem}
 This is again intuitive. If we treat both our slab and spike components similarly, the posterior will look very similar to the prior.

\begin{rem}
  If we let $\tau_1^2 \to \infty$ while keeping all else fixed, we have $p \to 0$.
\end{rem}
 
 This is an interesting result; there may exist some pathological choice of prior (i.e. very large $\tau_1$) due to which no value for $m,\sigma$ can indicate that the weight was coming from the slab component. This is counterintuitive, since - in its essence, our work is close to testing the significance of $m$ -- we should expect a higher value $m$ to recommend a higher value of $p$. Indeed, we reckon this is a specific case of Lindley's paradox \citep{lindley}. This can also be extended by taking $\tau_0^2 \to 0$ and keeping all else fixed, in which case, $p \to 1$.

\subsection{Quantifying Feature importance}

We consolidate the individual weight influences, denoted by \(\{p_i\}\), to estimate the importance of any node at any layer. This strategy is particularly useful for determining feature importance, as features correspond to the nodes in the first layer. We employ the `connection weights' method as proposed by \citet{olden2004accurate}, where we compute the multiplicative path probability for each path that begins with a specific feature and ends at the target, and then we average these path probabilities.

Formally, consider an $L$-layer Bayesian Neural Network (BNN), where each weight has an associated inclusion probability $p$. Let \(P^{(l)} \in \mathbb{R}^{n_{l} \times n_{l-1}}\) represent the node influence matrix between the $l$-th and $(l-1)$-st layers, where \([P^{(l)}]_{ij}\) is the inclusion probability for the weight connecting the $i$-th node in layer $l$ to the $j$-th node in layer $(l-1)\). Additionally, define $n_0 = p$ and \(n_{L+1} = 1\). The feature importance for the $j$-th input feature can then be defined as:

\begin{align}
    \psi_j =  \bigg[ \dfrac{1}{ \prod_{i= 1}^L n_i} \big( P^{(L+ 1)} P^{(L)} \cdots P^{(1)} \big)^T  \bigg]_j.  \label{eq:psi}
\end{align}

However, note that the estimated feature relevance values may become very small in practice. This is due to the fact that $\psi_j$ is a product of many probabilities, and thus the absolute values may be of little meaning.
To alleviate this issue, we will further define a scaled version of the feature importance which will take value from $0$ to $1$:
\begin{align}
    \phi_j = \frac{\psi_j - \displaystyle{\min_k} (\psi_k) }{\displaystyle{\max_k}(\psi_k) - \displaystyle{\min_k} (\psi_k) }. \label{eq:phi}
\end{align}

\section{Experiments}
\label{sec:exp}

In this section, we motivate the use of sparsity to compress our models. Experiments 1 to 3 are dedicated to the aforementioned feature selection strategies. Through a comprehensive set of simulation settings, we test our method to combine individual weight relevances into a measure of feature relevance and illustrate the feature selection algorithm based on this measure. These experiments provide significant evidence that removing unnecessary elements from the model greatly enhances its generalizability. Experiments 4 to 9 provide evidence for the efficacy of the weight pruning strategy. In Experiment 4, we demonstrate that we can retain the predictive power of a trained network while significantly reducing its complexity by pruning unnecessary weights, thus making the network much lighter. Building on this, Experiment 5 extends the pruning methods to real-world regression datasets, following the experimental setup outlined by \citet{Lobato2015}. Our results across ten datasets show that the full model achieves comparable or better errors in all cases, with the pruned model (50\% reduction) often performing comparably or better than other methods. However, some datasets highlight the importance of a larger model size. In Experiments 6 and 7, we demonstrate application on computer vision tasks. We extend comparison with experiments ran in Bayes by Backprop (BbB) on the MNIST and the CIFAR-10 datasets \citep{blundell2015weight,krizhevsky2009learning}, showing that our method incurs a significantly lower accuracy drop under extreme pruning conditions. Experiment 8 extends our method's application to Convolutional Neural Networks, where it performs on par with competing approaches, surpassing the Horseshoe method in some cases \citep{louizos2017bayesian}. Finally, in Experiment 9, we demonstrate the scalability of our method on a large VGG-like network \citep{vgglikecifar10blog}, achieving comparable or better error rates with significantly higher sparsity compared to the Horseshoe methods \citep{louizos2017bayesian}.

\subsection{Experiments on Simulated Datasets}

\subsubsection{Experiment 1: Checking Relationship Between Influence and $\psi$ } \label{sec:ex1}
First, let us consider the simplest situation. We have a two-variable linear model to test how the estimated feature relevance varies when the actual relevance of the features is varied. 

We generate $n = 2000$ copies of $X_1, X_2, \epsilon \overset{iid}{\sim} N(0,1)$ and define 
$$y_i = (1 - \alpha) x_{i1} + \alpha {x_{i2}} + \epsilon_i,$$
where $\alpha = \{0, 0.05, \dots, 1\}$. We fit a two-hidden-layer BNN 
with $20$ and $10$ hidden nodes. 

Once we have the fitted model, the predicted relative contributions of $X_1, X_2$ can be obtained  by finding $\psi_1, \psi_2$ according to \eqref{eq:psi}. To provide validity to our measures, we compare the predicted relative feature importances, to the actual relative contribution  of $X_2$ on $Y$. This is denoted by $I$, and defined as 
\begin{align*}
     I &= 1 - \dfrac{\sum\left(y_i - \alpha x_{i2} \right)^2}{\sum y_i^2}.
\end{align*} 

\begin{center}
    \begin{figure}[H]
        \centering
        \includegraphics[scale = 0.35]{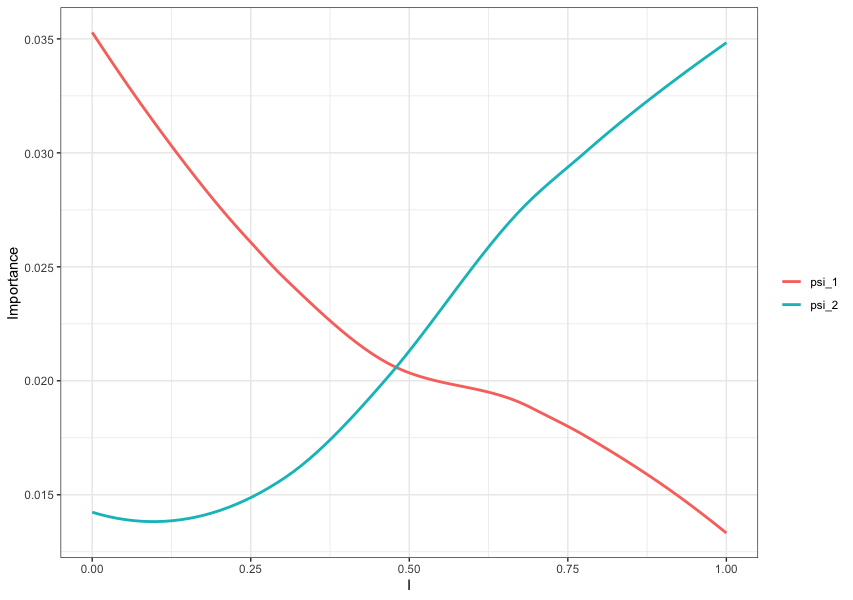}
        \caption{$I$ vs $\psi_1, \psi_2$ (Experiment 1). As the importance of $X_2$ goes from $0$ to $1$, its estimated relevance goes up while the relative importance of $X_1$ goes down simultaneously.}
        \label{fig:I v psi}
    \end{figure}
\end{center}

$I$ is chosen as a scale-free measure of the actual feature relevance. As we move $ \alpha$ from $0$ to $1$, we make $X_2$ more relevant than $X_1$ -- as a result, $I$ also rises from $0$ to $1$. Since the objective of this experiment is to compare relative contributions, and the network is small enough, we chose to use $\psi$ \eqref{eq:psi}. However, moving forward, for more general cases, we shall only use $\phi$ \eqref{eq:phi} measures.   
We observe the relationship between $I,\psi_1, \psi_2$ in Fig.  (\ref{fig:I v psi}). Per expectation, $I$ and $\psi_2$ share an almost linear increasing relationship, and $I$ and $\psi_1$ share an almost linear decreasing relationship. Thus, we can be confident that our estimate of the relative feature importance is able to catch the relative influence that feature has well enough. 
 

\subsubsection{Experiment 2: Generalizing for Many Input Features}
\label{sec:ex2}
We can further generalize Experiment 1 for many variables with more complicated data-generating functions and with varying effect sizes for each feature. 

We generate $n = 2000$ copies of $X_1, X_2, \dots, X_{D}, \epsilon \overset{iid}{\sim} N(0,1)$ and define
\begin{align*}
    y_i = \sum_j f(X_{ij})\beta_j + \epsilon_i.  
\end{align*}
Here, $\beta_j = \frac{j}{\alpha}$, 
$j  = 1, 2, \dots, D$.
The hyperparameters are as follows: $D$ is the total number of features being used, $\alpha$ varies the overall level of signal-to-noise in the data by controlling $\beta$. We use several different $\alpha = \{1, 2, ..., 10\}$. And finally, $f$ determines the relation between features and the response. We use two different such functions for illustration, one linear and one non-linear:
$$f(x)  = x,  \text{ or } f(x) = e^{|x|} - 2x + \sin(2\pi x).
$$

\begin{center}
    \begin{figure}
        \centering
        \includegraphics[scale = 0.22]{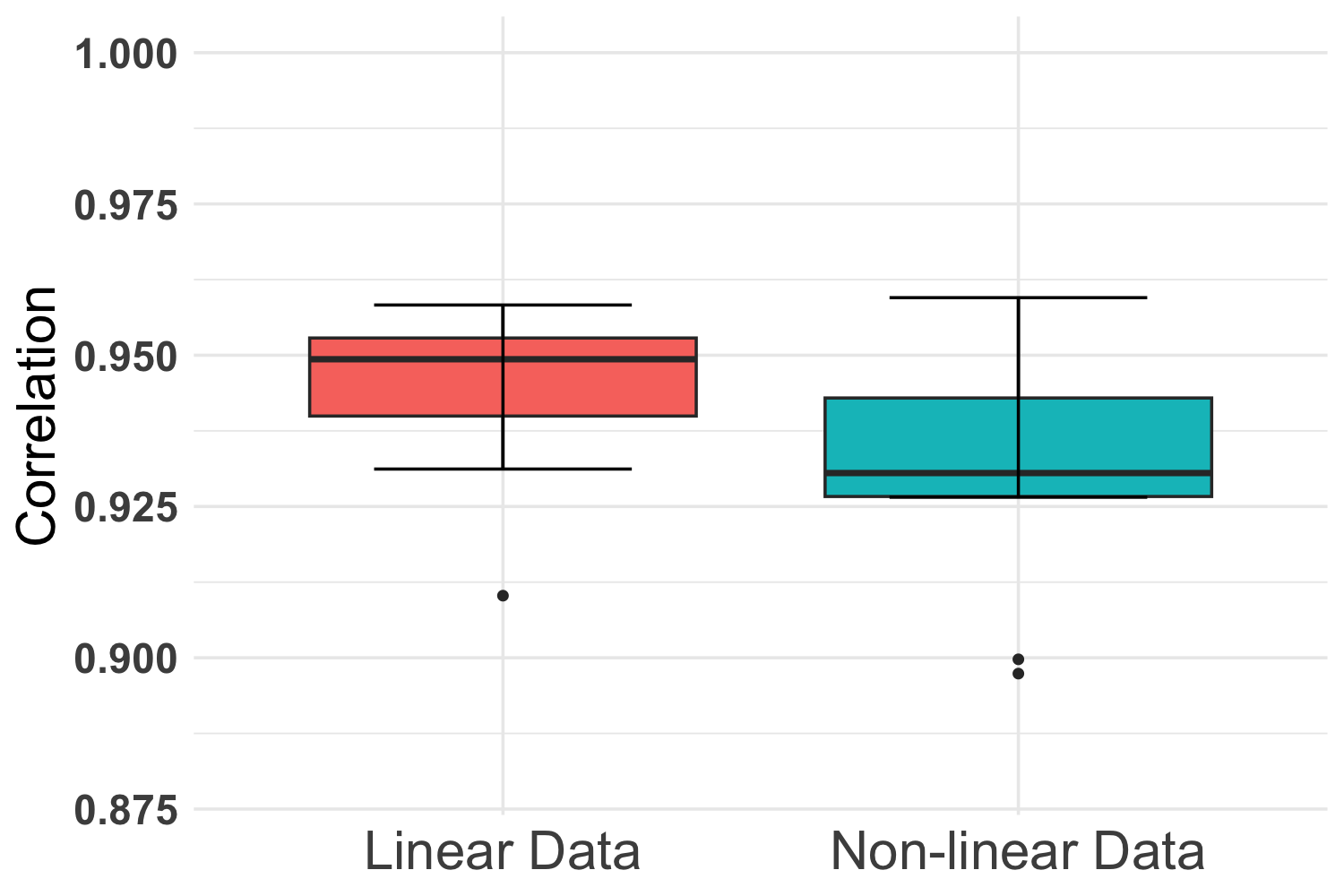}
        \caption{Correlation between Actual and Estimated Feature Importance. In all cases, correlations are high -- however, the linear data-generating processes have lower variability.}
        \label{fig:exp2}
    \end{figure}
\end{center}

The purpose of this experiment is to inspect the correlation between $\phi_j$ and $\beta_j$ with the understanding that the latter is a monotone function of the actual importance of the $j^{th}$ feature and the former is an estimate of it. We observe the relationship in Figure \ref{fig:exp2}. We can verify the correlation is within the range $(0.9, 0.95)$ for all datasets, which further strengthens our belief that the measure $\phi$ captures the relative distribution of the feature importance accurately enough. We also note that relationships are stronger and more stable in the case of linear data-generating models.

\subsubsection{Typical Values of $\phi_j$}

Let us take this opportunity to discuss what the typical values of $\phi_j$ would look like. As mentioned earlier, using $\psi_j$ directly might be problematic since it is a product of $p_i \in (0, 1)$ and hence may become too small (e.g. for higher depth) to work with effectively. In contrast, $\phi_j$ represents the relative importance of each feature and is always valued between $0$ and $1$. Higher values of $\phi_j$ indicate greater influence of the feature.

\begin{center}
    \begin{figure}
        \centering
        \includegraphics[scale = 0.18]{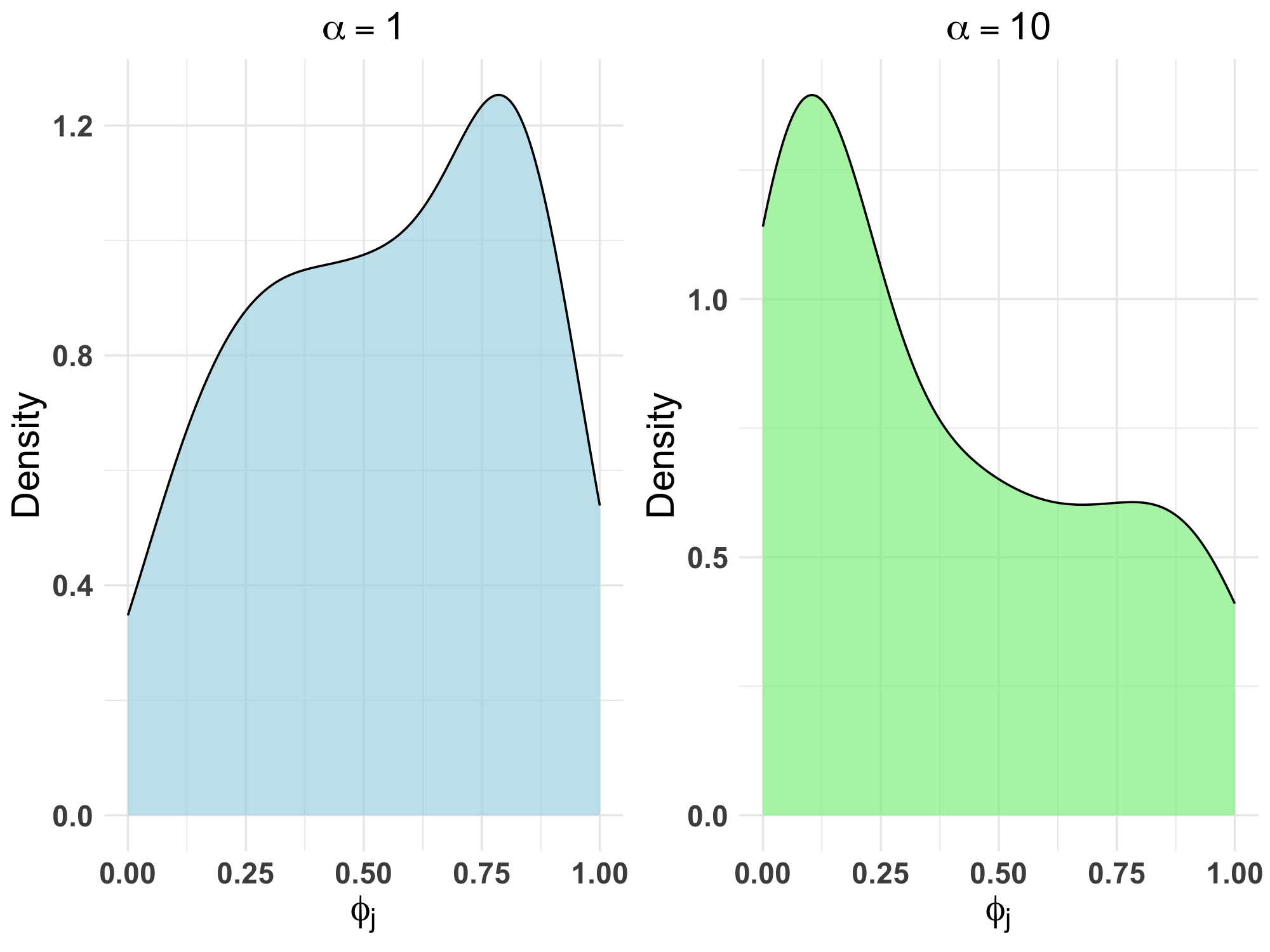}
        \caption{Correlation between Actual and Estimated Feature Importance. In all cases, correlations are high -- however, the linear data-generating processes have lower variability.}
        \label{fig:exp2_a}
    \end{figure}
\end{center}

Figure \ref{fig:exp2_a} illustrates two cases from this experiment. We are still operating in the non-linear data regime described by:
\[
y_i = \sum_j f(X_{ij}) \beta_j + \epsilon_i
\]
where $\beta_j = \frac{j}{\alpha}$. As shown in Figure \ref{fig:exp2}, $\phi_j$ aligns well with the relative contribution of each feature $\beta_j$ through correlation analysis. Furthermore, for any fixed $j$, a higher value of $\alpha$ corresponds to a lower signal-to-noise ratio, leading to less influence. This effect is observable in Figure \ref{fig:exp2_a}: for smaller $\alpha$, the density of $\phi_j$ is left-skewed, indicating a greater number of features with high influence. Conversely, for larger $\alpha$, the density is right-skewed, suggesting that fewer features have a high influence.

This behavior will be leveraged in the next section, where we will use $\phi_j$ to select a smaller set of active features by dropping less significant ones.

\subsubsection{Experiment 3: Checking Prediction Power}

Previous experiments provide ample evidence in support of the statement that our estimated feature relevance can efficiently capture the relative feature importance structure inherent in the dataset, even for complex data-generating processes. We will use this fact to do feature selection and demonstrate that the prediction power is improved by using a smaller BNN instead of a more complicated one. We will also use several other models for baseline comparisons.

Every dataset in consideration will have $n = 2000$ observations. Of which $80\% $ are used for training and the rest for testing. Each dataset will consist of $D$ features $X_1, X_2, ..., X_D \overset{iid}{\sim} N(0,1)$. Associated with every feature is a \textit{sparsity indicator} $Z_1, Z_2, ..., Z_D \overset{iid}{\sim}\text{Bernoulli}(\pi)$ and an \textit{effect size coefficient} $\beta_j = \frac{j}{\alpha}$. The response variable $Y$ is generated as follows with $\epsilon_i \overset{iid}{\sim} N(0,1)$:
\begin{align}
    y_{i} = \sum_j f(X_{ij}) \beta_j Z_j + \epsilon_i. \label{eq:exp3}
\end{align}

The hyperparameters of these experiments are $\alpha, f, D, \pi$. 
The new and  most important one among these is the sparsity hyperparameter 
$\pi$. This dictates what proportion of the features are actually active via determining how many $Z_i$s will be non-zero in any dataset. Rest are as defined in Experiment 2

The focal part of this experiment is testing the predictive and selective power of the feature selection method. For every dataset, we fit a BNN using our objective function. 
Based on the trained parameters, we determine the relative contribution of each feature according to \eqref{eq:psi}. Then we make a smaller model by only keeping the 'important' features in the model and discarding the rest. Determining whether to keep a feature or not is done based on a threshold. Initially, we show the use of a fixed $80 \%$ quartile threshold, i.e. we only keep features with top $20 \%$ feature relevance. Later on, we will demonstrate the shortcomings of this method and how to find a dynamic threshold by employing cross-validation. The formal process is as follows:

\begin{algorithm}[H]
\caption{BNN with Variable Selection ($BNN_{VS}$) }
  \begin{algorithmic}[1]
  \Function{VS}{$\textbf{P} = \{P_1, ..., P_L\}$}
     \State{\bf Convert:}  $P \to \phi_j$ \hspace{0.1 cm} $\forall$ \hspace{0.1 cm} $j = 1, 2, \dots D$ 
     \State{\bf Compute:} $r = 80^{th} \text{ \%-ile of the } \phi_j$
     \State $\hat{Z}_j= \mathbb{I}(\phi_j \ge r)$ 
     \State $\mathcal{W}^T$ = $\left(W_1, W_2, \dots, W_D \right)$ 
     \State \textbf{return} $\mathcal{W}$
  \EndFunction
  \vspace{0.5 cm}
  \Function{fit}{$y, \textbf{X}, \mathcal{W}$}
     \State  $\textbf{X}_{vs} =  \textbf{X} \cdot diag (\mathcal{W})$
     \State{\bf Train:} BNN $y \sim g (\textbf{X}_{vs})$
  \EndFunction
  \end{algorithmic}
\end{algorithm}

\begin{center}
    \begin{figure}[H]
        \centering
        \includegraphics[scale = 0.16]{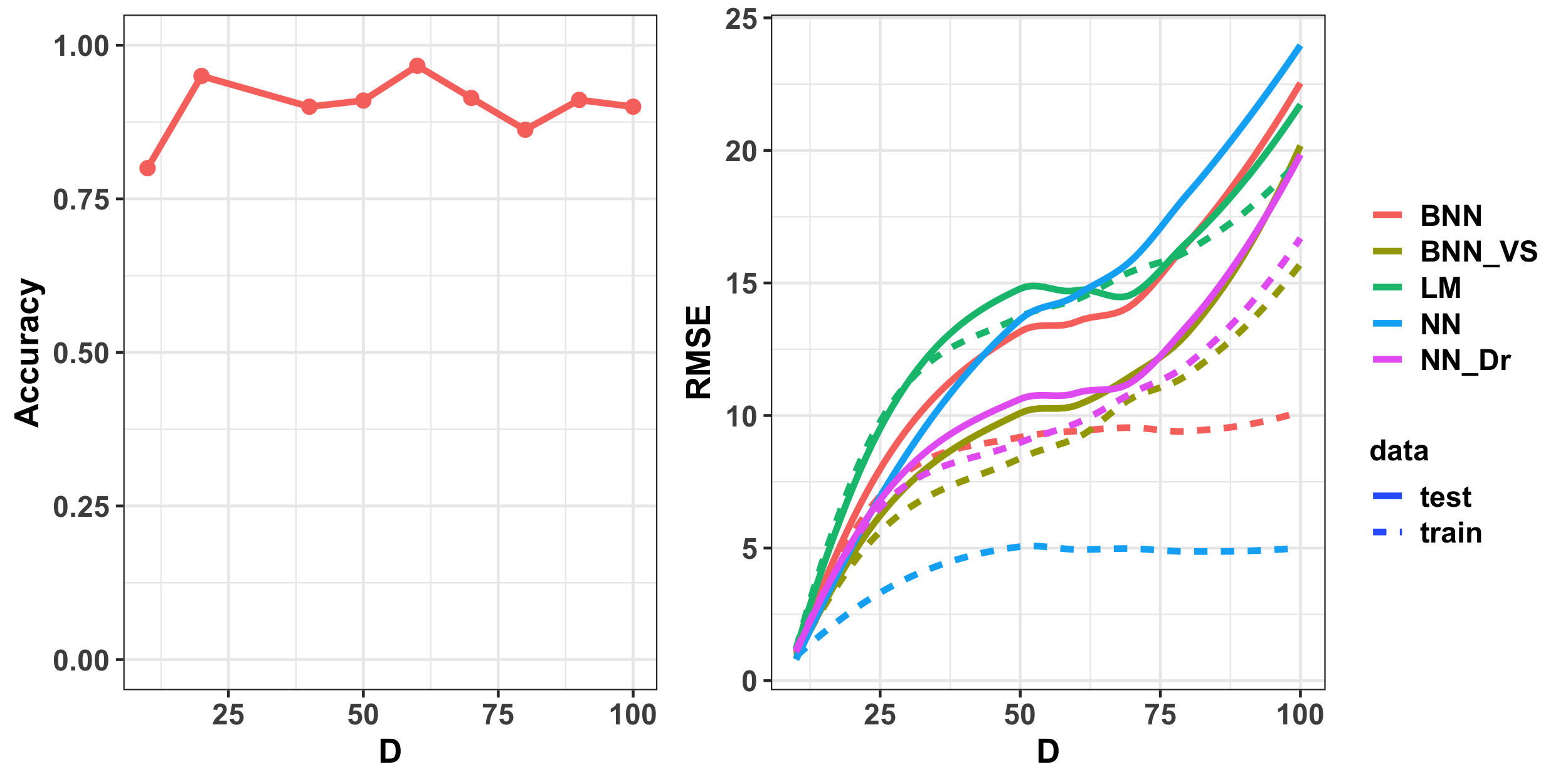}
        \caption{We vary $D$, the number of features. While keeping the following hyperparameters fixed: $\alpha = 2, f(x) = e^{|x|} -2x + \sin(2\pi x), \pi = 0.2$. We notice (Left) high selection accuracy and (Right)improved test errors in all such datasets. }
        \label{fig:exp3_1}
    \end{figure}
\end{center}

\begin{center}
    \begin{figure}[H]
        \centering
        \includegraphics[scale = 0.16]{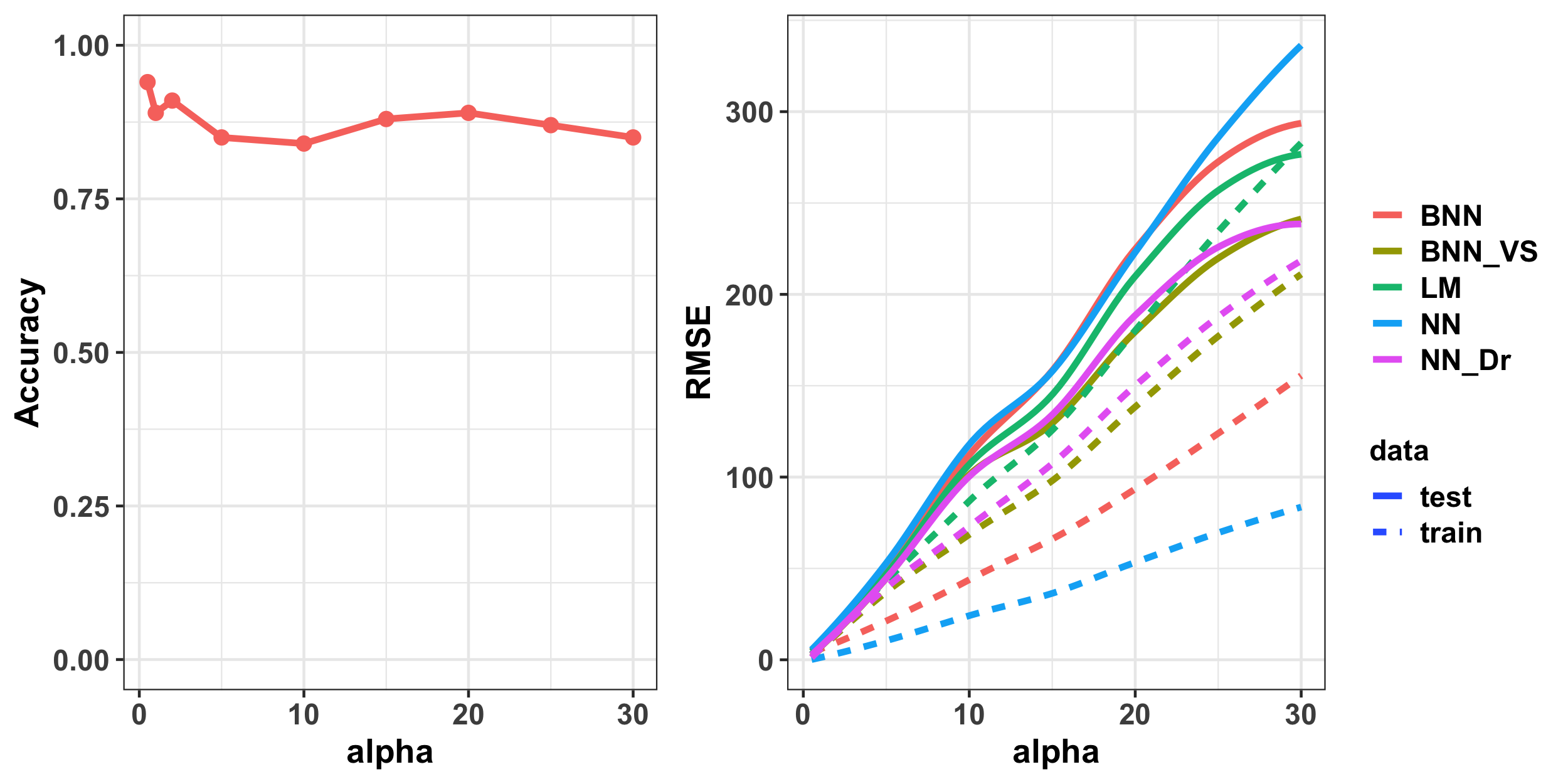}
        \caption{We vary $\alpha$, the signal to noise ratio indicator.  While keeping the following hyperparameters fixed fixed: $D = 100, f(x) = e^{|x|} -2x + \sin(2\pi x), \pi = 0.2$. We again notice (Left) high selection accuracy and (Right) improved test errors in all such datasets }
        \label{fig:exp3_2}
    \end{figure}
\end{center}

\begin{center}
    \begin{figure}[H]
        \centering
        \includegraphics[scale = 0.16]{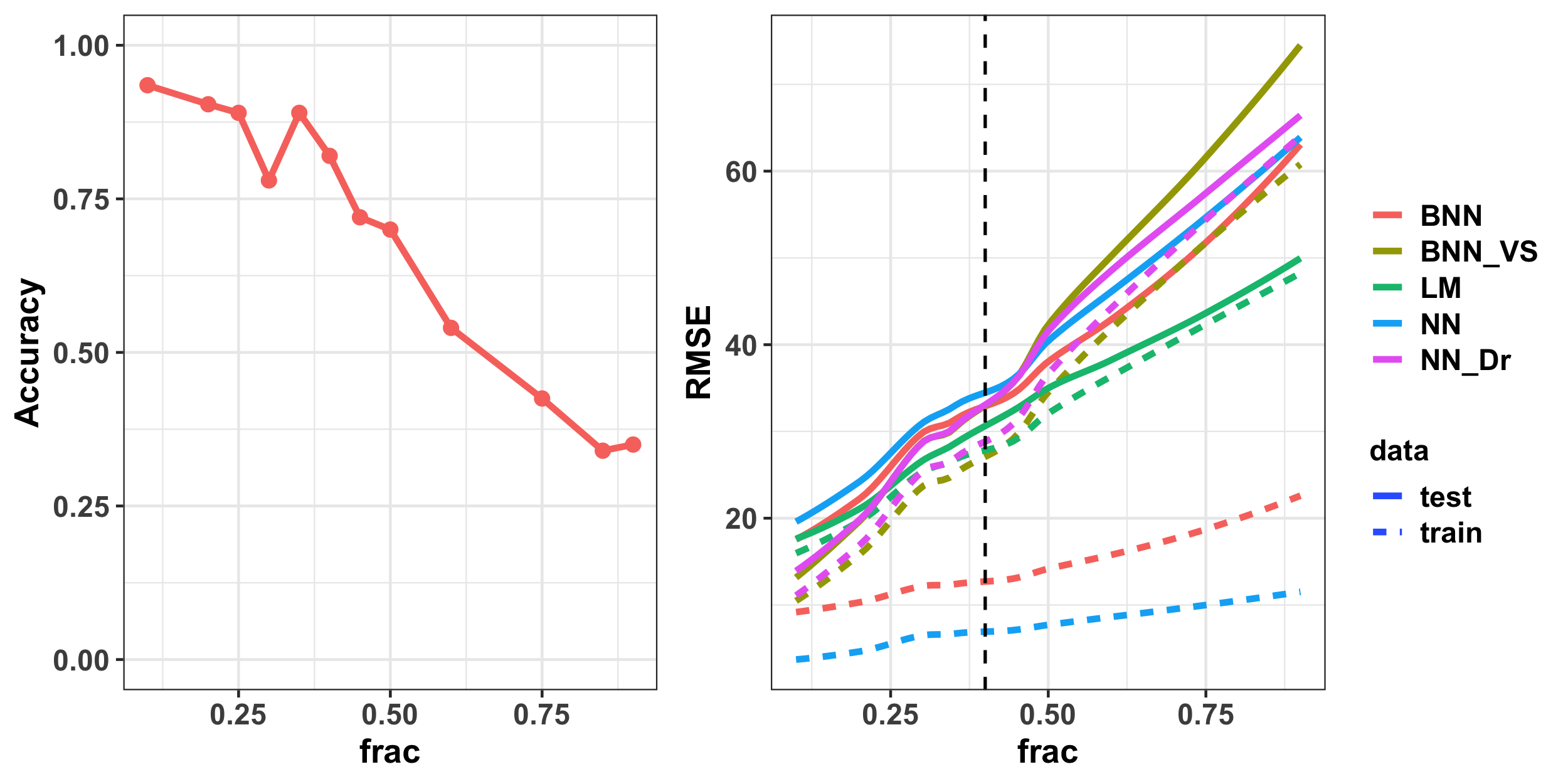}
        \caption{we vary the sparsity by varying $\pi$, while keeping the following hyperparameters fixed: $D = 100, f(x) = e^{|x|} -2x + \sin(2\pi x), \alpha = 2$. We notice (Left) a significant drop in accuracy and (Right) generalizability as more and more active features are ignored. }
        \label{fig:exp3_3}
    \end{figure}
\end{center}

The quantity $\hat{Z}_j$ for each feature is of utmost interest. This denotes the inclusion of any feature in our smaller model. Ideally, this will be identical to $Z_j$ for every feature. We can obtain the selection accuracy of the method for each dataset by verifying this relationship $
        Acc. = \dfrac{1}{D} \sum_1 ^ {D}  \mathbb{I}  (Z_j = \hat{Z}_j)
$
We also measure whether discarding inactive features provide us with better generalization by comparing train and test MSEs in each of the datasets. We compare the improvement of generalizability of our method with the same obtained from the following baseline algorithms, 
$NN$ (Unrestricted Neural Network), $NN_{Dr}$ (Neural Network with Dropout), $LM$ (Linear Model),  $BNN$ (Simple BNN)
All of the Neural Networks are trained on a 2 hidden layer architecture with 20, and 10 hidden nodes respectively. We vary the hyperparameters $\alpha, f, \pi, D$ to generate different datasets and compare outputs in Figures (\ref{fig:exp3_1}, \ref{fig:exp3_2}, \ref{fig:exp3_3}).

Firstly, note that we use only top $20\%$ features in the $BNN_{VS}$ model, which means that our entire model is shrunk to $27$\% of its original size. However, Figures (\ref{fig:exp3_1}, \ref{fig:exp3_2}) indicate that changing the effect size or the number of factors does not have any impact on the efficacy; we still get about a $90$\% accuracy in selecting the active features. Further, 
Figures  $3, 4$  also show the similarity between $BNN_{VS}$ and $NN_{Dr}$ methods, they both use shrinkage, in some form and thus demonstrate  similar reductions in test errors in several scenarios that largely eclipse other methods. The other methods serve as a yardstick, we notice $LM$ always has the highest training error and somewhat better test error showing underfit, while $NN$ has the least training error and worst test error signaling clear signs of overfit. 

Figure $5$ shows a very interesting dynamic. Here we intentionally introduce misspecification. We vary the sparsity indicator from $0 - 95 \%$, so each dataset has a different proportion of active features. But we only select top $20 \%$ of active features and discard the rest. We can notice that as the proportion of active features increases, the accuracy drops, as expected. $BNN_{VS}$ still performs best in terms of test error as long as the proportion of active features increases is less than $50\%$(middle column), however, as we make more features active, the misspecification catches up -- $BNN_{VS}$ show very high train as well as test error (rightmost column). Perhaps more interestingly, $NN_{Dr}$, which still takes all the features as input also suffers similarly.

    

\subsubsection{Cross Validation for Finding Optimal Thereshold}

\begin{center}
   \begin{figure}[h]
     \centering
     \begin{subfigure}[t]{0.475\textwidth}
         \centering
         \includegraphics[width=1 \textwidth]{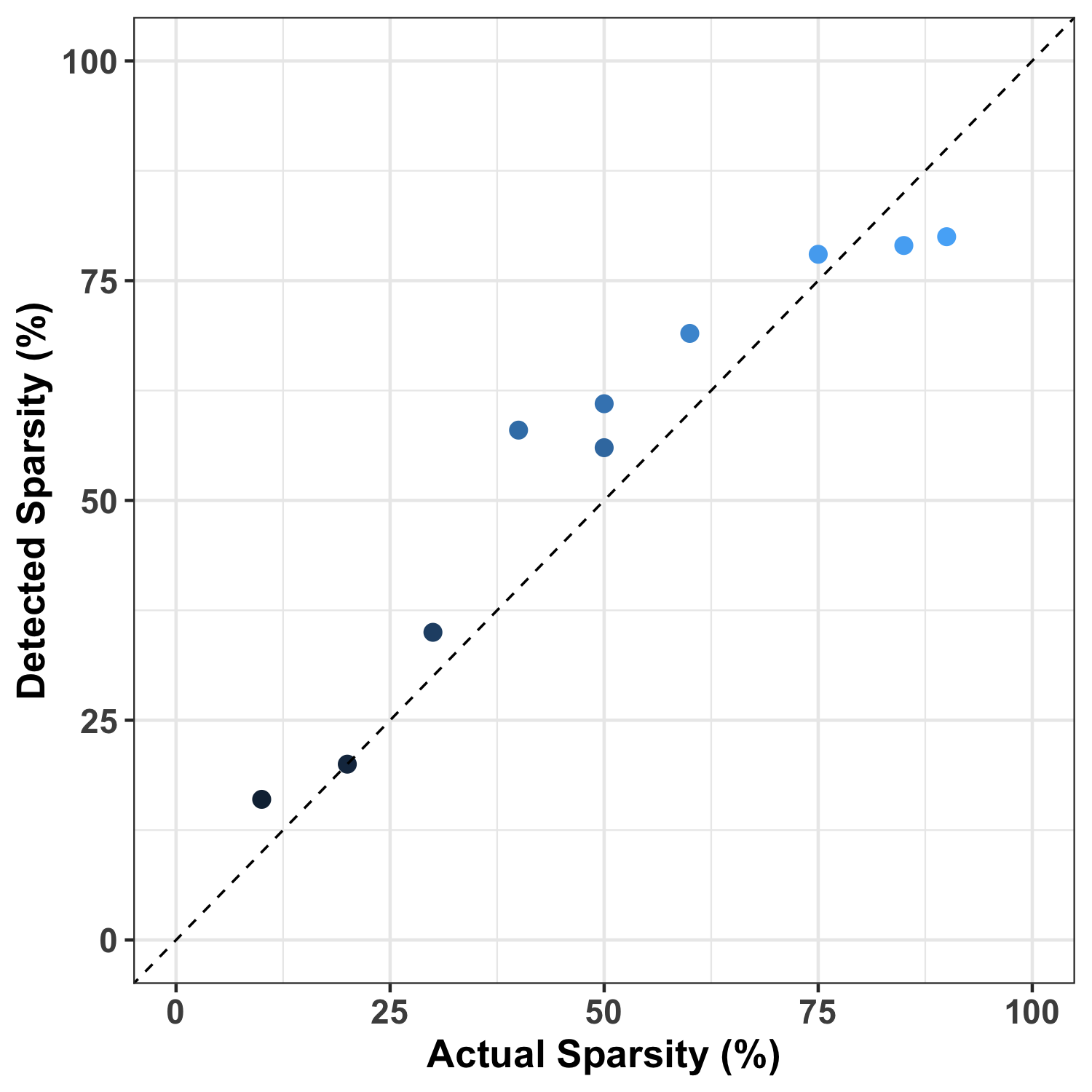}
         \caption{ Actual Sparsity vs Detected Threshold. Points on and around $y=x$ line indicate proper threshold detection.}
         \label{fig:exp4_1}
     \end{subfigure}
     \hfill 
     \begin{subfigure}[t]{0.475\textwidth}
         \centering
          \includegraphics[width=1 \textwidth]{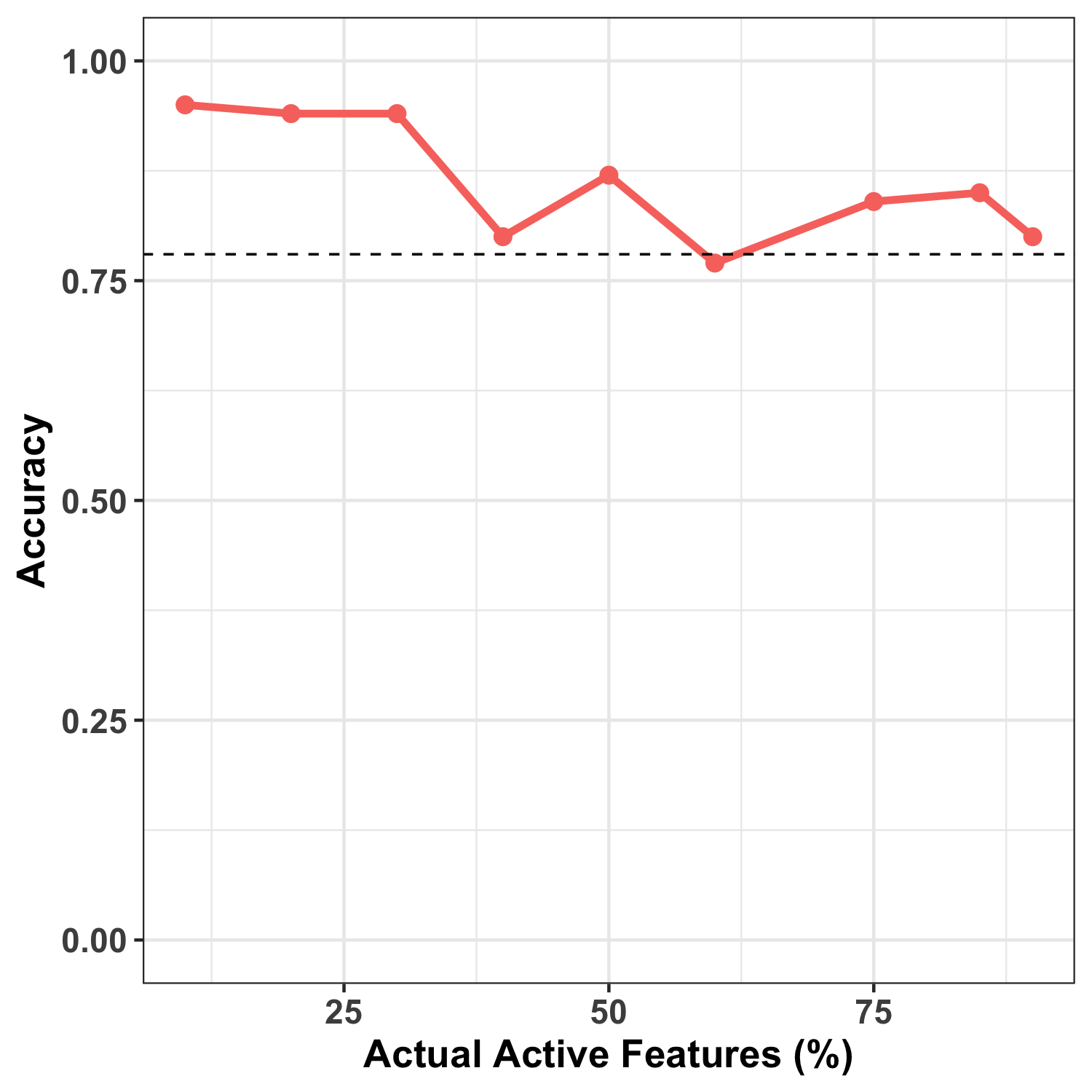}
         \caption{ Selection accuracy as the sparsity is varied. Accuracy never dips down and always stays above  $80 \%$. }
         \label{fig:exp4_2}
     \end{subfigure}
     \hfill
        \caption{ (a)We can detect feature selection threshold via CV. The actual proportions of active features align well with the detected proportions of active features. (b) Using such dynamically determined thresholds can help us find active features with high accuracy in any dataset with any given amount of sparsity.}
        \label{fig:three graphs}
\end{figure} 
\end{center}

\begin{center}
    \begin{figure}[h]
        \centering
        \includegraphics[scale = 0.18]{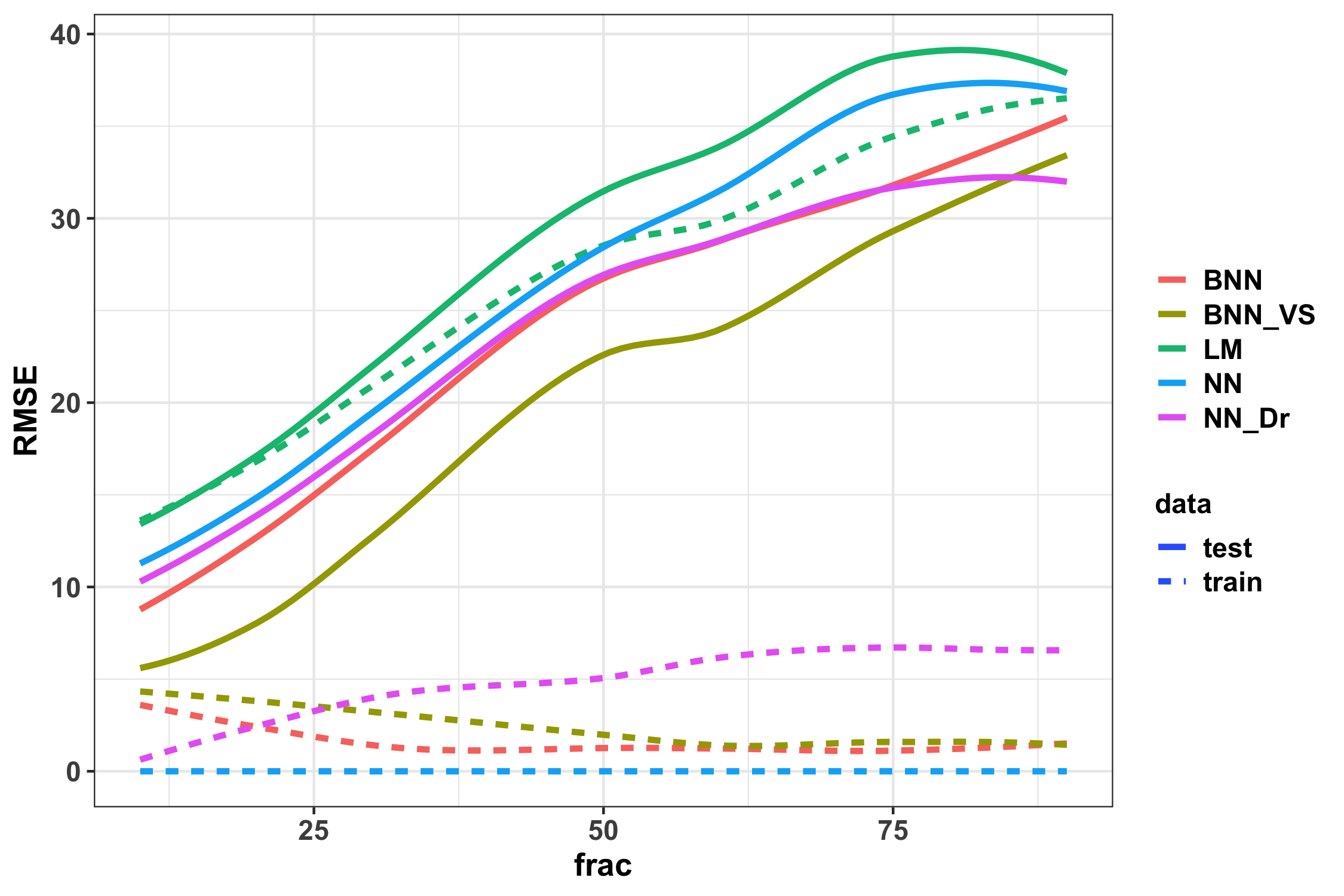}
        \caption{Capturing the correct features makes sure the smaller network almost always outperforms all other methods
        }
        \label{fig:exp4_3}
    \end{figure}
\end{center}

At this stage, we can employ cross-validation to understand what part of the input feature is indeed active. Let us focus on the data corresponding to Figure \ref{fig:exp3_3}, we have $100$ features, but the sparsity proportion is varying from $0-100 \%$
To automatically detect what proportion of the features are indeed active, we have implemented a 10-fold cross-validation setup. This dynamic threshold value obtained from CV can be used for feature selection instead of a pre-specified proportion. The results we find are insightful. If we repeat the experiments with varying proportions of active features, we can notice (Figure \ref{fig:exp4_1}) the estimated proportion of active features lines up relatively well with the actual proportion. More importantly, using this threshold now, we can select appropriate features (Figure \ref{fig:exp4_2}) -- we observe that the accuracy of selecting relevant features never dips down, even though we do not know in advance what proportion of the features is actually active. Selecting correct features with high accuracy gives us better (Figure \ref{fig:exp4_3}) test errors in almost all cases by some margin.  

\subsubsection{Experiment 4: Weight Pruning in a Trained Network} \label{subseq:exp4}

\begin{center}
    \begin{figure}[H]
        \centering
        \includegraphics[scale = 0.4]{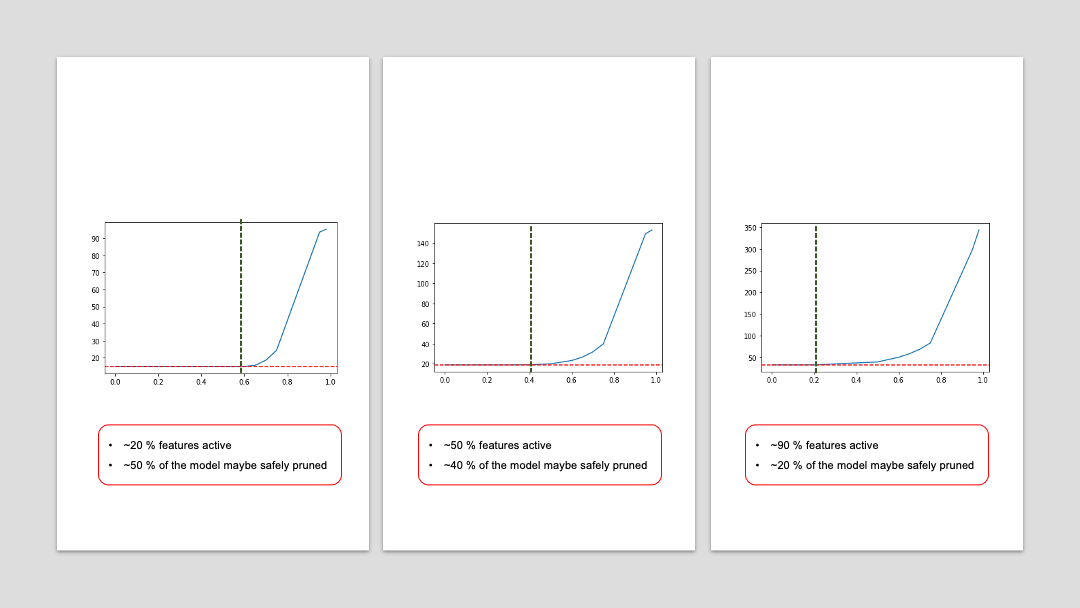}
        \caption{Weight Pruning in a Trained Network, (from left to right) more proportion of features are active. Thus, we will need to retain more proportion of the trained BNN to keep similar errors in test data.   }
        \label{fig:exp4}
    \end{figure}
\end{center}

For compressing an already trained neural network we can directly remove the weights that are not useful as determined by the final `$p$' value obtained as an outcome of Algorithm \ref{algo:1}. Since a higher value of `$p$'  indicates a stronger slab component and thus higher influence on the trained network. Using this strategy, we have the option to tune our level of compression based on our tolerance to error and the availability of resources. We demonstrate this again with the same data used in Experiment 3. If $\pi = 0.2$, i.e., only around $20\%$ of the features are active, we can safely prune around $50-60 \%$ of the whole network and retain practically the same precision. Moreover, using simulation, we can notice that the volume of this `sufficient' network changes when more features are active. If instead, we had $\pi = 0.5$, i.e. $50 \%$ of the features were actually active, we can only prune around $40 \%$ of the network before we start hemorrhaging precision. And when $90 \%$ of the features are active this number goes down to $20 \%$ - so about $80 \%$ of the model is useful. This phenomenon is illustrated in Figure \ref{fig:exp4}.

\subsection{Experiments on Regression Datasets}

\subsubsection{Experiment 5: Weight Pruning for Regression Analysis } \label{sec:prune_UCI}

In Section \ref{subseq:exp4}, we outlined the methodology for leveraging our algorithm's output to prune redundant weights from neural networks, resulting in significantly leaner networks while preserving their predictive capabilities. Extending this experiment to various real-world datasets, we align with the experimental setups described in \citet{Lobato2015}, \citet{ghosh2017model} by selecting ten publicly available datasets. The datasets are detailed in Table \ref{tab:UCI_datasets}. Through regression analysis conducted on these standard datasets with standard architecture, we aim to demonstrate the efficacy of our approach.

\begin{table}[h]
\centering
\captionsetup{skip=20pt} 
\caption{Dataset Characteristics}
\vspace{0.1 in}
\label{tab:dataset}
\begin{adjustbox}{scale  = 1}
\begin{tabular}{@{}l|c|c@{}}
\toprule
\textbf{Dataset} & \textbf{N} & \textbf{p} \\
\midrule
Boston Housing \citep{boston}& 506 & 13 \\
Concrete Compression Strength \citep{misc_concrete_compressive_strength_165} & 1030 & 8 \\
Energy Efficiency \citep{misc_energy_efficiency_242} & 768 & 8 \\
kin8nm & 8192 & 8 \\
Naval Propulsion \citep{misc_condition_based_maintenance_of_naval_propulsion_plants_316} & 11934 & 16 \\
Combined Cycle Power Plant \citep{misc_combined_cycle_power_plant_294} & 9568 & 4 \\
Protein Structure \citep{misc_physicochemical_properties_of_protein_tertiary_structure_265}& 45730 & 9 \\
Wine Quality Red \citep{misc_wine_quality_186} & 1599 & 11 \\
Yacht Hydrodynamics \citep{misc_yacht_hydrodynamics_243}& 308 & 6 \\
Year Prediction MSD \citep{misc_year_prediction_msd_203} & 515345 & 90 \\
\bottomrule
\end{tabular}
\label{tab:UCI_datasets}
\end{adjustbox}
\end{table}

The experimental setup maintains consistency with prior works, ensuring comparability with methodologies employed for comparison purposes. In this setup, each dataset undergoes a standard splitting procedure throughout our experiments into training and testing sets, with $90\%$ of the data allocated for training and the remaining $10\%$ for testing. The datasets are also normalized so that the input features and the response have zero mean and unit variance in the training set. The normalization on the response is omitted during prediction.  

For training, we employed the Algorithm as described in section \ref{sec:meth2} on an `oversized' neural network with one hidden layers consisting of $50$ nodes each. However, for the \texttt{year} dataset, we augment the network with 100 hidden nodes. The performance evaluation metric utilized is the test root mean square error (RMSE) and the  hyperparameters \{$\log\tau_1, \log\tau_0, \pi$\} were set to $\{1, -6, 0.5\}$ respectively.

The resultant test RMSEs are summarized in Table \ref{tab:rmse}. Here, Droprate denotes the proportion of removed weight. We examine how predictive performance changes as more and more weights are set to zero. The table unveils varying levels of redundancy within the original network for different datasets. For example, certain datasets (e.g., \texttt{wine}) demonstrate the networks' ability to maintain predictive power even with nearly $90\%$ of their weights pruned. In contrast, others (e.g., \texttt{Power Plant, Protein}) require the retention of most (more than $80\%$) of the network's weights to preserve comparable accuracy. This discrepancy is expected, reflecting the variability in the inherent complexity of the data generation processes.

\begin{center}
    \begin{figure}[H]
        \includegraphics[scale = 0.2]{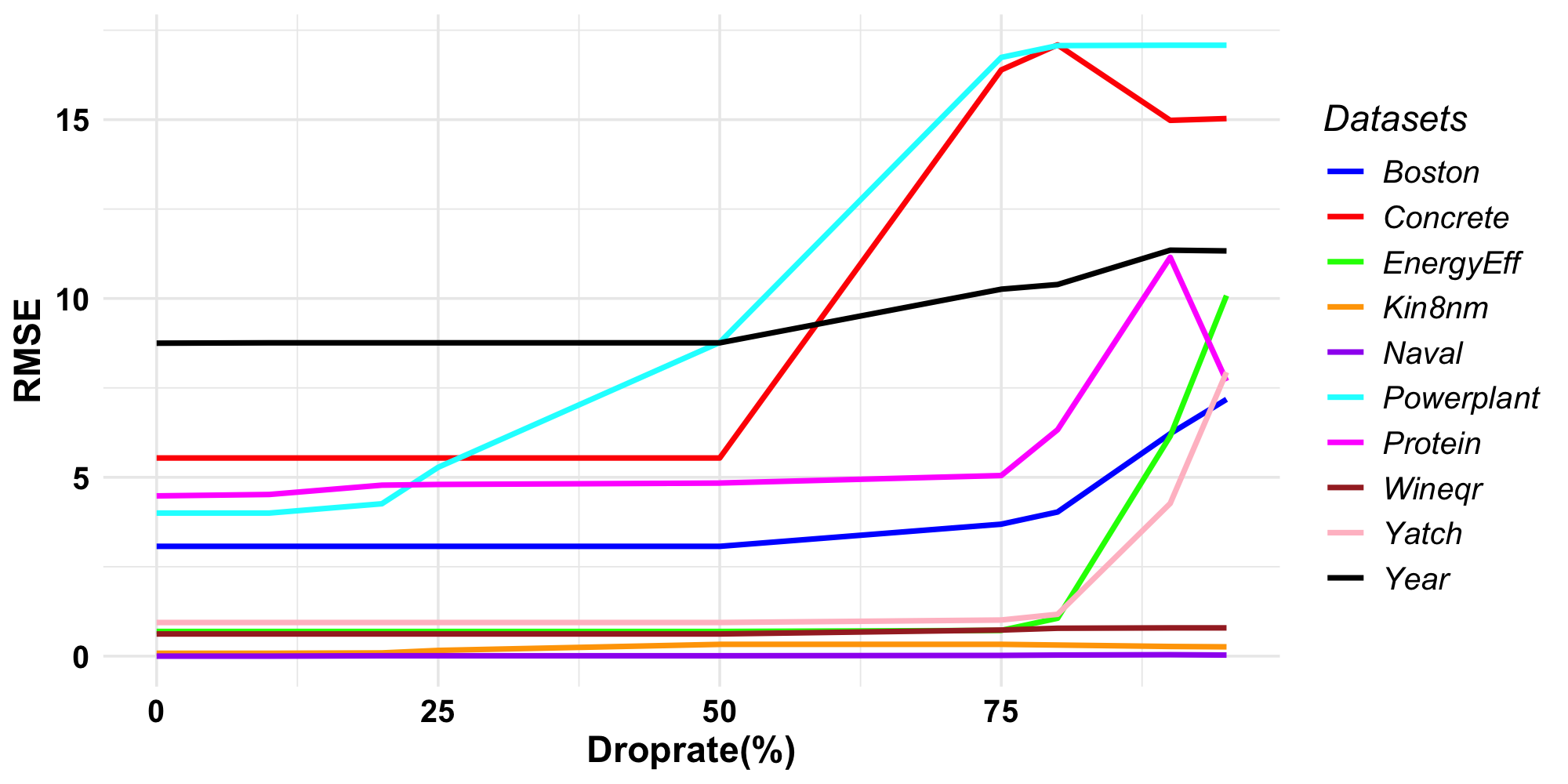}
        \caption{}
        \label{fig:Rmse_UCI}
    \end{figure}
\end{center}

This behavior is illustrated in Figure \ref{fig:Rmse_UCI}, which depicts the trajectory of test RMSE values as more unimportant weights are pruned. In most datasets, we observe consistent performance until approximately all but $50\%$ of the weights are removed. However, for datasets such as \texttt{Powerplant} and \texttt{Protein}, the drop in performance occurs much earlier. We can further posit that employing a larger initial network would likely reveal higher levels of redundancy, as will be demonstrated in later experiments.

\begin{table}[htbp]
  \centering
  \caption{Mean RMSE values for different datasets at varying Droprates}
  \label{tab:rmse}
  \begin{adjustbox}{scale = 0.65} 
    \begin{tabular}{c|c|c|c|c|c|c|c|c|c|c}
      \toprule
      \textbf{DR (\%)} & \textbf{Boston} & \textbf{Concrete} & \textbf{EnergyEff} & \textbf{Kin8nm} & \textbf{Naval} & \textbf{Powerplant} & \textbf{Protein} & \textbf{Wineqr} & \textbf{Yatch} & \textbf{Year} \\
      \midrule
      0    & 3.07  & 5.54 & 0.69  & 0.08  & 0.00  & 4.00  & 4.48  & 0.62  & 0.94 & 8.75 \\
      10   & 3.07  & 5.54 & 0.69  & 0.08  & 0.00  & 4.00  & 4.52  & 0.62  & 0.94 & 8.76 \\
      20   & 3.07  & 5.54 & 0.69  & 0.09  & 0.01  & 4.26  & 4.78  & 0.62  & 0.94 & 8.76 \\
      25   & 3.07  & 5.54 & 0.69  & 0.16  & 0.01  & 5.28  & 4.80  & 0.62  & 0.94 & 8.76 \\
      50   & 3.07  & 5.54 & 0.69  & 0.33  & 0.01  & 8.77  & 4.84  & 0.62  & 0.94 & 8.76 \\
      75   & 3.69  & 16.39 & 0.72 & 0.33  & 0.02  & 16.74 & 5.05  & 0.73  & 1.01 & 10.26 \\
      80   & 4.03  & 17.09 & 1.06 & 0.31  & 0.03  & 17.07 & 6.33  & 0.78  & 1.17 & 10.39 \\
      90   & 6.22  & 14.98 & 6.15 & 0.27  & 0.04  & 17.08 & 11.15 & 0.79  & 4.27 & 11.35 \\
      95   & 7.18  & 15.03 & 10.08& 0.26  & 0.03  & 17.08 & 7.69  & 0.79  & 7.94 & 11.33 \\
      \bottomrule
    \end{tabular}
  \end{adjustbox}
\end{table}

Next, we will assess the predictive efficacy of our model against some of the most prominent methods in the Bayesian Neural Network field that have demonstrated identical setups. Alongside the results for the full model (sBNN-50), we will also include results for the model with 50\% of the weights pruned (sBNN-50). This will help us verify not only the efficacy of the model itself but also of the pruning process. We will compare these results with those from four comparable models. Probabilistic Backpropagation (PBP) \citep{Lobato2015} serves as our baseline, this paper laid down the foundation for this experiment setup and along with \citet{blundell2015weight, graves2011} was among the pioneers of the modern implementation of the BNN. Additionally, we will include the Dropout \citep{gal2016dropout} method, as our method is intrinsically related to it and may be seen as a more generalized version of it. Further, we have added two more recent approaches that are closely related to our work, both of these employ some form of shrinkage prior.  Horse-shoe BNN (HS-BNN) \citep{ghosh2017model} adopts a horseshoe prior and Sparse Variational BNN (SVBNN) \citep{bai2020efficient} uses a different form of spike and slab prior (a mass at zero instead of a continuous spike prior). All the results are taken from their respective papers, the results for \citet{bai2020efficient} are taken from \citet{liu2024implicit}.
 
We find that the full model demonstrates competitive performance across all datasets, achieving the lowest mean RMSE in 7 out of 10 datasets. Among these, in \texttt{Energy}, \texttt{Yacht}, and \texttt{Year}, we observe substantial improvement over existing comparable methods. For the remaining three datasets, the error bars (computed as two standard errors from the mean value in each direction) overlap, indicating comparable results. Encouragingly, the model with a 50\% reduction in load either outperforms or shows comparable performance to other methods in many datasets. However, exceptions are noted in the \texttt{Power Plant}, \texttt{kin8nm}, and \texttt{Protein} datasets, where the importance of having a larger model size becomes evident.

{ \small
\begin{table}[htbp]
  \centering
  \captionsetup{justification=centering} 
  \caption{Comparison of Different Methods on Various Datasets}
  \label{tab:rmse_Comp}
  \begin{adjustbox}{scale = 0.82}
    \begin{tabular}{l|c|c|c|c|c|c}
      \toprule
      \textbf{Dataset} & \textbf{sBNN} & \textbf{sBNN-50} & \textbf{SVBNN} & \textbf{HS-BNN} & \textbf{Dropout} & \textbf{PBP} \\
      \midrule
      Boston    & 3.07 $\pm$ 0.14 & 3.07 $\pm$ 0.14 & 3.17 $\pm$ 0.58 & 3.32 $\pm$ 0.66 & \textbf{2.97 $\pm$ 0.19} & 3.01 $\pm$ 0.18 \\
      Concrete  & 5.54 $\pm$ 0.14 & 5.54 $\pm$ 0.14 & 5.57 $\pm$ 0.47 & 5.66 $\pm$ 0.41  & \textbf{5.23 $\pm$ 0.12} & 5.67 $\pm$ 0.09 \\
      Energy    & \textbf{0.69 $\pm$ 0.08} & \textbf{0.69 $\pm$ 0.08} & 1.92 $\pm$ 0.19 & 1.99 $\pm$ 0.34 & 1.66 $\pm$ 0.04 & 1.80 $\pm$ 0.05 \\
      Kin8nm    & \textbf{0.08 $\pm$ 0.00} & 0.33 $\pm$ 0.02 & 0.09 $\pm$ 0.00 & 0.08 $\pm$ 0.00 & 0.10 $\pm$ 0.00 & 0.10 $\pm$ 0.00 \\
      Naval     & \textbf{0.00 $\pm$ 0.00} & 0.01 $\pm$ 0.00 & 0.00 $\pm$ 0.00 & 0.00 $\pm$ 0.00  & 0.01 $\pm$ 0.00 & 0.01 $\pm$ 0.00 \\
      Power Plant & \textbf{4.00 $\pm$ 0.04} & 8.77 $\pm$ 0.49 & 4.01 $\pm$ 0.02 & 4.03 $\pm$ 0.15  & 4.02 $\pm$ 0.04 & 4.12 $\pm$ 0.04 \\
      Protein   & 4.48 $\pm$ 0.02 & 4.84 $\pm$ 0.02 & \textbf{4.30 $\pm$ 0.05} & 4.39 $\pm$ 0.04  & 4.36 $\pm$ 0.01 & 4.73 $\pm$ 0.01 \\
      Wine      &\textbf{0.62 $\pm$ 0.01} & \textbf{0.62 $\pm$ 0.01} & 0.62 $\pm$ 0.04 & 0.63 $\pm$ 0.04  & 0.62 $\pm$ 0.01 & 0.64 $\pm$ 0.01 \\
      Yacht     & \textbf{0.94 $\pm$ 0.06} & 0.94 $\pm$ 0.06 & 1.10 $\pm$ 0.27 & 1.58 $\pm$ 0.23  & 1.11 $\pm$ 0.09 & 1.02 $\pm$ 0.05 \\
      Year      & \textbf{8.75 $\pm$ NA} & 8.76 $\pm$ NA & 8.87 $\pm$ NA & 9.26 $\pm$ 0.00  & 8.85 $\pm$ NA & 8.88 $\pm$ NA\\
      \bottomrule
    \end{tabular}
  \end{adjustbox}
\end{table}
}




\subsection{Experiments on Image Classification Datasets Using Fully Connected Neural Networks}

\subsubsection{Experiment 6: Classification on MNIST}
In this section, we apply our method to MNIST dataset in comparison with the experimental results of Section 5.1 from \cite{blundell2015weight}. To make fair comparisons, we used the same network structures and preprocessing procedure from \cite{blundell2015weight}. Specifically, we trained various networks of two hidden layers of rectified linear units and a softmax output layer with 10 units on the MNIST digits dataset, consisting of 60000 training and 10000 testing images. Each pixel of every image is divided by 126 before feeding to the network. For training, we ran SGD on our objective function for 300 epochs and considered learning rate of 0.05, 0.01 and 0.005 with minibatches of size 128. For hyperparameters tuning, we consider $\pi \in \{0.25, 0.5, 0.75\}$, $\log(\tau_1) = -1 + c$, $\log(\tau_0) = -5 - c$, where $c \in [0, 3]$, and perform binary search to determine c. Hyperparameters were chosen directly based on test accuracy, no validation set was considered. The chosen hyperparameters are reported in Table \ref{tab:MNIST_test_error}


Similar to other variational Bayesian frameworks, our objective function \eqref{eq:solution} is amenable to minibatch optimization, a strategy commonly used for neural networks. In every epoch of optimization, the training data $\mathcal{D}$ is split randomly into a partition of M equally sized subsets, $\mathcal{D}_1, \cdots, \mathcal{D}_M$. Each gradient is then averaged within one of these minibatches. Proposed by \cite{graves2011}, a minibatch cost function can be defined by injecting one batch of data $\mathcal{D}_i$ on the loss term while dividing the penalty term with $M$ as follows,
\begin{equation}
    \label{eq:minibatch_objective_even}
  \mathcal{J}_i(\theta) =  -\mathbb{E}_{q(\mathbf{W})} \log p(\mathcal{D}_i | \mathbf{W}) + \frac{1}{M}\mathcal{R}(\theta ).  
\end{equation}
such that $\sum_{i = 1}^M \mathcal{J}_i(\theta) = \mathcal{J}(\theta)$. Minibatch cost function \eqref{eq:minibatch_objective_even} evenly distribute the penalty across all minibatches. In this experiment, we use a different minibatch cost definition with non-uniformly distributed penalty suggested by \cite{blundell2015weight} as follows,
\begin{equation}
    \label{eq:minibatch_objective}
  \mathcal{J}_i(\theta) =  -\mathbb{E}_{q(\mathbf{W})} \log p(\mathcal{D}_i | \mathbf{W}) + r_i\mathcal{R}(\theta )
\end{equation}
where $\sum_{i=1}^M r_i = 1$. \cite{blundell2015weight} suggests that minibatch cost \eqref{eq:minibatch_objective} with schema $r_i = \frac{2^{M-i}}{2^M - 1}$ works well. To put the comparison on equal footing, we used this same minibatch schema.

We start with comparing our result with \cite{blundell2015weight} by reporting the test error. In Table \ref{tab:MNIST_test_error}, our method's classification error rate is comparable to that of the methods reported in Blundell's work.

\begin{table}[H]
\centering
\caption{Classification Error Rates on MNIST}
\begin{adjustbox}{max width= \textwidth}
\begin{tabular}{c|c|c|c|l}
\toprule
\multicolumn{1}{c|}{Method} & \multicolumn{1}{c|}{\# Units/Layer} & \multicolumn{1}{c|}{\# Weights} & \multicolumn{1}{c|}{Test Error} & \multicolumn{1}{c}{Hyperparameters} \\
\midrule
\multirow{3}{*}{sBNN} & 400 & 500k & 1.45\% & $\pi=0.5$, $\log(\tau_1)=1.0$, $\log(\tau_0)=0.002$ \\ 
& 800 & 1.3m & 1.47\% & $\pi=0.5$, $\log(\tau_1)=1.648$, $\log(\tau_0)=0.002$ \\ 
& 1200 & 2.4m & 1.58\% & $\pi=0.5$, $\log(\tau_1)=0.375$, $\log(\tau_0)=0.007$ \\ 
\bottomrule
\end{tabular}
\end{adjustbox}
\label{tab:MNIST_test_error}
\end{table}

In Table \ref{tab:MNIST_weight_pruning}, we examine the effect of replacing the variational posterior on some of the weights with a constant zero. We took the network with two layer of 1200 units trained in table \ref{tab:MNIST_test_error}, and order the weights by their posterior inclusion probability. We removed the weights with the lowest inclusion probability. The model demonstrates strong performance even when $98\%$ of its weights are removed. We also provide the corresponding results for the method proposed in \citet{blundell2015weight}. It is essential to emphasize that our primary goal is to compare different sparsity levels within the same method rather than comparing between different methods.

Applying the heuristic signal-to-noise approach used in \cite{blundell2015weight}, their model experiences a reduction in efficacy of $7.75\%$ when $98\%$ of the neurons are removed. In contrast, our model only experiences a $3.1\%$ decrease in efficacy under the same conditions. This contrast is consistently observed in every other sparsity condition as well. 
\begin{table}[H]
\centering
\caption{Classification Errors after Weight Pruning}
\begin{adjustbox}{max width=\textwidth}
\begin{tabular}{c|c|c|c}
\toprule
Proportion Removed & \# Weights & Test Error (sBNN) & Test Error (BbB) \\
\midrule
0\% & 2.4m & 1.58\% & 1.24\% \\ 
50\% & 1.2m & 1.57\% & 1.24\% \\ 
75\% & 600k & 1.52\% & 1.24\% \\ 
95\% & 120k & 1.54\% & 1.29\% \\ 
98\% & 48k & 1.61\% & 1.39\% \\ 
\bottomrule
\end{tabular}
\end{adjustbox}
\label{tab:MNIST_weight_pruning}
\end{table}

\begin{center}
    \begin{figure}[H]
        \centering
        \includegraphics[scale = 0.8]{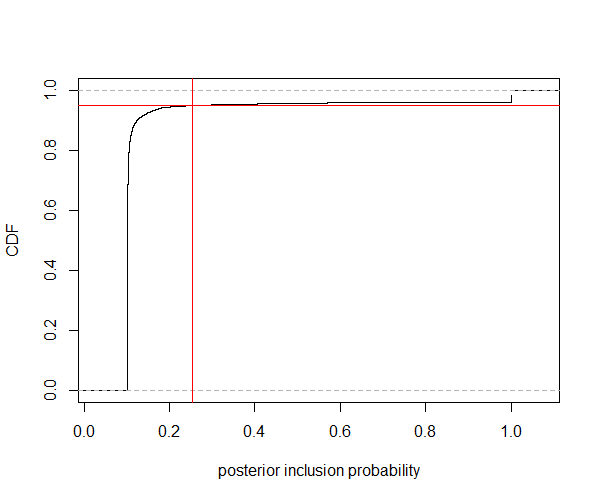}
        \caption{CDF of posterior inclusion probability over all weights in the network. The red line denotes the 95\% cut-off.}
        \label{fig:MNIST_post_inc_dist}
    \end{figure}
\end{center}

In Figure \ref{fig:MNIST_post_inc_dist}, we examine the distribution of posterior inclusion probability of the network in Table \ref{tab:MNIST_weight_pruning}. Most of the posterior inclusion probabilities are concentrated around 1 or smaller than 0.25, there are more than $95\%$ of weights that have a posterior inclusion probability less than 0.25.

\subsubsection{Experiment 7: Classification on CIFAR-10}
In their original paper, \citet{blundell2015weight} only performed the aforementioned experiment with the MNIST dataset. Here, we extend the same methodology to the larger CIFAR-10 dataset. Unlike MNIST, CIFAR-10 is not greyscale; it consists of images with three color channels and dimensions of $32 \times 32$ pixels for each channel, with a 10-class classification output. For this experiment, we flatten all input pixels and train a two-hidden layer neural network, with each hidden layer containing 1200 nodes. Additionally, we train a model using BbB with the same architecture for comparison.

In summary, the results obtained from our experiments on the CIFAR-10 dataset reaffirm the efficacy of weight pruning techniques in reducing model complexity without significantly sacrificing classification performance. We observe a gradual increase in test error rates as the proportion of pruned weights increases, reflecting the loss of representational capacity in the model. However, even with significant weight pruning (e.g., $98\%$ of weights removed), the performance degradation remains manageable, particularly considering the substantial reduction in model size and computational requirements.

Comparison with \citet{blundell2015weight}'s method reveals similar trends to those observed in MNIST. The error rate in the original model is similar, and they remain comparable until $90 \%$ of weights are pruned. However, beyond that, our model shows greater robustness. For instance, under high pruning regimes, such as when $98 \%$ of the parameters are dropped, our model experiences a $29.3\%$ drop in accuracy from the full model, whereas BbB suffers a drastic $66 \%$ drop. This example further emphasizes that if large pruning is the goal with spike and slab prior, the second moment of the weights (as in our method) is more appropriate than the signal-to-noise ratio (as in BbB). The details of this experiment are presented in table \ref{tab:CIFAR_weight_pruning}.

\begin{table}[H]
\centering
\caption{Classification Errors after Weight Pruning in CIFAR10 Dataset}
\begin{adjustbox}{max width=\textwidth}
\begin{tabular}{c|c|c|c}
\toprule
Proportion Removed & \# Weights & Test Error (sBNN) & Test Error (BbB) \\
\midrule
0\% & 5.14m & 26.34\% & 25.81\% \\ 
50\% & 2.57m & 27.02\% & 25.86\% \\ 
75\% & 1.29k & 27.04\% & 25.82\% \\ 
90\% & 514k & 26.63\% & 26.26\% \\ 
95\% & 257k & 27.45\% & 29.64\% \\ 
98\% & 103k & 33.92\% & 42.83\% \\ 
\bottomrule
\end{tabular}
\end{adjustbox}
\label{tab:CIFAR_weight_pruning}
\end{table}

\subsection{Experiments and Sparsity Comparison on Image Classification Datasets Involving Convolutional Layers }
\label{sec:lenet}

\subsubsection{Experiment 8: Classification on MNIST using Lenet structures}
In this segment, we further demonstrate the ability of our algorithm to induce sparsity in classification tasks and compare it with several widely used algorithms. For this purpose, as directed in \cite{blalock2020state}, we will apply our method to train sparse neural networks on the MNIST dataset using a fully- connected architecture LeNet-300-100 and a convolutional architecture LeNet-5 \citep{Lenet-300-100}. These networks were trained from a random initialization and without data augmentation. We also extended our method to convolutional networks by associating all parameters from convolutional layers with spike and slab prior and learn the variational distribution in the exact same way we did for parameters in fully connected layer. During pruning phase, the cutoff threshold applies to variational inclusion probability of parameters from all convolutional layers and fully connected layers, so that sparsity also occur within the convolutional layer during pruning. Formally, we define sparsity using the following formula: $$ Sparsity =1 - \frac{|w \ne 0|}{|w|}, $$
where $|w \ne 0|$ denotes the count of non-zero weights and $|w|$ represents the total number of weights in the network. Higher sparsity denotes a leaner network,  goal is to achieve a similar error range with as high sparsity as possible. This definition is similar to but not exactly same as the ones used in \citet{louizos2017bayesian, molchanov2017variational}.

To contextualize our findings, we provide a comparative analysis, including error and sparsity metrics, against several state-of-the-art algorithms. These include Deep Compression (DC) \citep{Han2016DeepCC}, Dynamic Network Surgery (DNS) \citep{guo2016DNS} and Soft Weight Sharing (SWS) \citep{ullrich2017soft}, Sparse Variational Dropout (SVD) \citep{molchanov2017variational}, and Bayesian Compression using Group Horseshoe (BC-GHS) \citep{louizos2017bayesian}. In Table \ref{tab:lenet_comparison}, we can see our method performs head to head with all the competing approaches, beating the widely used horseshoe method in both tasks. 

\begin{table}[H]
    \centering
    \begin{adjustbox}{width=0.8\textwidth}
    \begin{tabular}{c|p{3.5cm}|c|c}
        \toprule
        \textbf{Network} & \textbf{\emph{Method}} & \textbf{Error \%} & \textbf{Sparsity (\%)} \\
        \midrule
        \multirow{6}{*}{Lenet-300-100} & Original & 1.64 & - \\
         & DNS & 1.99 & 98.2 \\
         & SWS & 1.94 & 95.7 \\
         & Sparse VD & 1.92 & 98.5 \\
         & DC & 1.59 & 92.0 \\
         & BC-GHS & 1.80 & 89.4 \\
         & sBNN (This paper) & 1.94 & 95.0 \\
        \midrule
        \multirow{7}{*}{Lenet-5-caffee} & Original & 0.80 & - \\
         & DNS & 0.91 & 99.1 \\
         & SWS & 0.97 & 99.5 \\
         & Sparse VD & 0.75 & 99.3 \\
         & DC & 0.77 & 91.7 \\
         & BC-GHS & 1.01 & 99.4 \\
         & sBNN (This paper)  & 1.06 & 99.2\\
        \bottomrule
    \end{tabular}
    \end{adjustbox}
    \caption{Comparison of Error \% and Sparsity for different methods on Lenet architectures.}
    \label{tab:lenet_comparison}
\end{table}

\subsubsection{Discussion on Hyperparameter Tuning }

In deep-learning architecture, hyperparameter tuning is a crucial aspect of optimizing model performance. Specifically, our setup involves three key hyperparameters: $\pi$ (prior sparsity), $\tau_0$ (slab prior standard deviation), and $\tau_1$ (spike prior standard deviation). Initial experiments on smaller networks, as detailed in Section \ref{sec:prune_UCI}, indicated minimal discernible effects from these hyperparameters. However, a more extensive exploration in Section \ref{sec:lenet}, which focuses on larger networks, reveals a more nuanced impact on both sparsity and test error.

In summary, when using two closely spaced values of $(\log \tau_1, \log \tau_0)$, we observe a relatively higher test error in the full network but the network retains substantial efficacy post-pruning. On the other hand, selecting more distant values for $(\log \tau_1, \log \tau_0)$ tends to yield a lower test error for the full network. However, this configuration may lead to a rapid loss of network efficacy after at a lower pruning level. Essentially, this denotes a trade-off between  initial test error and the robustness of the network to pruning

\begin{center}
    \begin{figure}[H]
        \includegraphics[scale = 0.19]{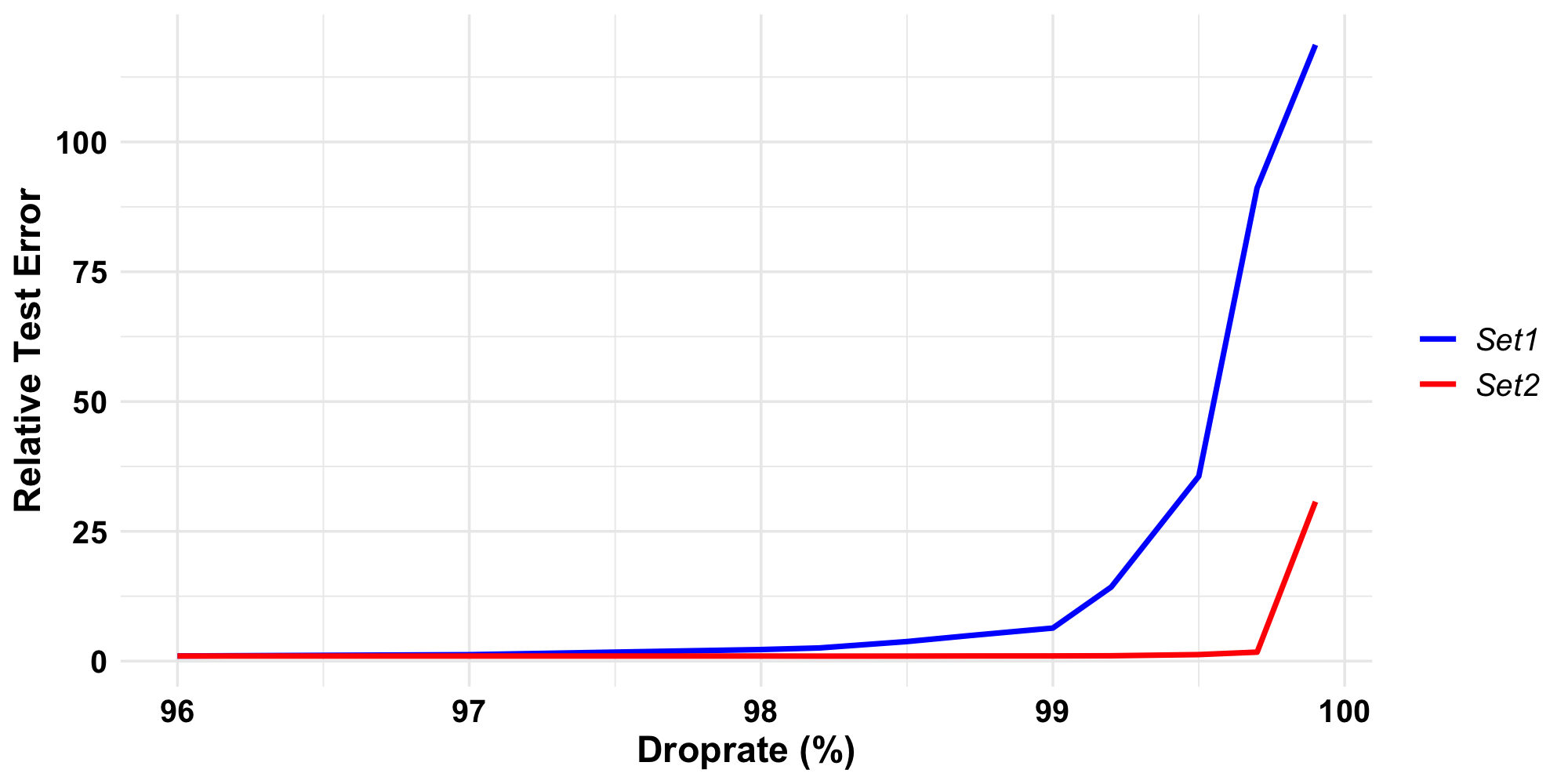}
        \caption{ We demonstrate the proportional increase in test error in the two scenarios, considering pruning levels beyond $96 \%$. Each line in the figure represents the test error at that level of pruning over the original test error for that model.}
        \label{fig:hyperparameter}
    \end{figure}
\end{center}

Let us demonstrate this behavior in action with the Lenet5 architecture on MNIST. In the table below we exhibit some musings with two sets of values of the hyperparameters. In set 1, we used two values of $(\log \tau_1 = 0, \log \tau_0 =-5)$ that are relatively farther. While in set 2, we used two relatively closer values of $(\log \tau_1 = -1, \log \tau_0 =-3)$. Our analysis,  as demonstrated in table \ref{tab:hyperparameter} and figure \ref{fig:hyperparameter}, shows that the choice of $(\tau_1, \tau_0)$ values significantly influences the performance of the network in achieving an optimal trade-off between test RMSE and network sparsity.

\begin{table}[htbp]
\centering
\begin{adjustbox}{max width=\textwidth}
\begin{tabular}{c|c|c}
\toprule
\textbf{Droprate} & \textbf{Test Error (Set 1)} & \textbf{Test Error (Set 2)} \\ 
\midrule
0.0 & 0.70 & 1.03 \\ 
\midrule
96.0 & 0.69 & 1.01 \\ 
97.0 & 0.87 & 1.01 \\ 
98.0 & 1.56 & 1.01 \\ 
98.2 & 1.78 & 0.99 \\ 
98.5 & 2.64 & 1.00 \\ 
98.7 & 3.39 & 1.02 \\ 
99.0 & 4.46 & 1.03 \\ 
99.2 & 9.99 & 1.06 \\ 
99.5 & 24.93 & 1.30 \\ 
99.7 & 63.82 & 1.78 \\ 
99.9 & 83.08 & 31.62 \\ 
\bottomrule
\end{tabular}
\end{adjustbox}
\caption{Test Error for Different Droprates}
\label{tab:hyperparameter}
\end{table}

As we can see, the test error with the full network is quite low in Set 1 ($0.7\%$). In fact, Set 1 achieves state-of-the-art error for this architecture with the full model but only maintains this performance up to $96\%$ pruning. Beyond that, the error rate increases rapidly, making this pruning level less comparable to other methods in Table \ref{tab:lenet_comparison}. 

In Set 2, the full model's error might be slightly higher ($1.03\%$), but more of the network can be pruned without significant efficacy loss. In this particular case, we can achieve $99.2\%$ sparsity with a $1.06\%$ test error, which is more comparable to methods such as BC-GHS \citep{louizos2017bayesian}. This result is demonstrated in Table \ref{tab:lenet_comparison}. Even beyond that, a $99.7\%$ pruning retains an error rate of $1.78\%$, within a percentage point of the original result, which one might find acceptable. For Set 1, similar accuracy is only achieved around $98.2\%$ sparsity and not beyond that. Figure \ref{fig:hyperparameter} illustrates this robustness (or lack thereof) by showing the proportional increase in test error compared to the original network. As can be observed, the difference between the two models becomes noticeable around the $97\%$ pruning stage and is highly significant when more than $98\%$ of the respective networks are pruned.

These findings underscore the importance of tuning the appropriate values for $\tau_1$ and $\tau_0$ to achieve a delicate balance between overall error and sparsity metrics. One might prioritize a slightly higher potential test error with a smaller network or aim for a lower test error while preserving more of the original network. Ultimately, determining the optimal middle ground hinges on the user's sensitivity to the trade-off between error mitigation and resource utilization.

\subsubsection{Experiment 9: Classification on CIFAR-10 with VGG-like structures}

In this section, we demonstrate the scalability of our method to larger network structures by applying it to a VGG-like network \citep{vgglikecifar10blog} on the CIFAR-10 dataset \citep{krizhevsky2009learning}. The VGG network is a sizable mode. comprising 13 convolutional layers and two fully connected layers, with each layer followed by batch normalization. This architecture totals approximately 15 million parameters, a size that has proven imperative in breaking the 90\% mark in test accuracy on CIFAR-10. However, such a large model is also ripe for pruning. We apply our pruning method to all weights in the convolutional and fully connected layers, removing all dropout layers from the architecture and using only our method for inducing sparsity.

Similar to the training and pruning strategy described in previous sections, we tuned our method's hyperparameters on a grid of values: $\pi \in \{0.25, 0.5, 0.75\}$, $\log(\tau_1) \in \{-6, -5, -4\}$, and $\log(\tau_0) \in \{-2, -1, 0\}$, and pruned the network based on the rank of variational inclusion probability. We trained our network with pre-trained weights and achieved a test accuracy of $91.4\%$ with the hyperparameters set at $\pi = 0.5$, $\log(\tau_1) = 0$, and $\log(\tau_0) = -5$. Sparsity analysis showed that approximately $98.5\%$ of the weights could be set to zero without significant change in accuracy.

As noted by \citet{blalock2020state}, the VGG-like architecture is notorious for non-comparability among works. To provide a direct comparison, we utilized the two Horseshoe Methods \citep{louizos2017bayesian} in table \ref{tab:VGG_comparison}. Our approach achieves a better or equivalent error rate at a much higher level of sparsity compared to these methods.

\begin{table}[H]
    \centering
    \begin{adjustbox}{width=0.8\textwidth}
    \begin{tabular}{c|p{3.5cm}|c|c}
        \toprule
        \textbf{Network} & \textbf{\emph{Method}} & \textbf{Error \%} & \textbf{Sparsity (\%)} \\
        \midrule
        \multirow{4}{*}{VGG-like} & Original & 7.6 & - \\
         & BC-GNJ & 8.6 & 93.3 \\
         & BC-GHS & 9.0 & 94.5 \\
         & sBNN (This paper) & 8.7 & 98.5 \\
        \bottomrule
    \end{tabular}
    \end{adjustbox}
    \caption{Comparison of Error \% and Sparsity for different methods on VGG architectures on CIFAR-10 dataset.}
    \label{tab:VGG_comparison}
\end{table}


\section{Conclusion}\label{sec:conclusion}

In this work, we developed a novel strategy to incorporate Bayesian methods for sparsity detection into deep learning frameworks. By utilizing the inclusion probabilities from a Bayesian Neural Network (BNN), induced by applying a spike-and-slab prior, we can effectively identify sparsity within the model. This identified sparsity can then be leveraged to compress the model significantly, reducing its size while retaining its predictive power.

Several intriguing future directions arise from this work. One potential avenue is to use the order of sparsity obtained via inclusion probabilities $p_i$ to hypothesize and determine the optimal architecture for the model, potentially leading to more efficient and tailored network designs. Additionally, exploring the theoretical guarantees of the feature selection methods we have proposed presents a rich area for further research. Understanding these guarantees could enhance the robustness and applicability of our methods.

We encourage practitioners in the Bayesian deep learning community to adopt this technique and \textit{illuminate their black-boxes} by applying it to larger and more complex projects. This approach not only offers practical benefits in terms of model compression and interpretability but also contributes to advancing the field of Bayesian deep learning by providing clearer insights into model structure and feature importance.
\newpage

\appendix{
\label{sec:suppA} 
\section*{S1: Proof for Proposition \ref{lemma:J:two-parts}}

Recall the objective function
\begin{align*}
    \mathcal{J}(\theta, \mathbf{p})  &= - \mathbb{E}_{q_{\theta}(\mathbf{W})} \mathbb{E}_{q_{\mathbf{p}}(\mathbf{Z})}  \log \frac{p(\mathcal{D} | \mathbf{W}) \cdot \pi(\mathbf{W}, \mathbf{Z})}{q_{\theta} (\mathbf{W}) \cdot q_{\mathbf{p}}(\mathbf{Z})}  \\
   & = -\mathbb{E}_{q_{\theta}(\mathbf{W})} \log p(\mathcal{D} | \mathbf{W}) - \mathbb{E}_{q_{\theta}(\mathbf{W})} \mathbb{E}_{q_{\mathbf{p}}(\mathbf{Z})} \log \frac{ \pi(\mathbf{W}, \mathbf{Z})}{q_{\theta} (\mathbf{W}) \cdot q_{\mathbf{p}}(\mathbf{Z})}.
\end{align*}
Write the second term as
\begin{align*}
  \text{(II)} & := \mathbb{E}_{q_{\theta}(\mathbf{W})} \mathbb{E}_{q_{\mathbf{p}}(\mathbf{Z})} \log \frac{ \pi(\mathbf{W}, \mathbf{Z})}{q_{\theta} (\mathbf{W}) \cdot q_{\mathbf{p}}(\mathbf{Z})} \\
    & = \sum_i \mathbb{E}_{q_{\theta}(\mathbf{W})} \left( \mathbb{E}_{q_{\mathbf{p}} (\mathbf{Z})} \left(  \log  \pi(W_i, Z_i) - \log q(Z_i) \right)  - \log q(W_i) \right).
\end{align*}
Calculating these terms individually, 
\begin{align*}
  &  \mathbb{E}_{q_{\mathbf{p}} (\mathbf{Z})} \left(\log  \pi(W_i, Z_i) - \log q(Z_i) \right) \\
    &= \mathbb{E}_{q_{\mathbf{p}} (\mathbf{Z})} \left(Z_i \left( \log \frac{\pi}{p_i}  + \log N(w_i; 0, \tau_1^2 ) \right) + (1 - Z_i)\left( \log \frac{1- \pi}{1 - p_i}  + \log N(w_i; 0, \tau_0^2 ) \right) \right) \\
    &= p_i \left( \log \frac{\pi}{p_i}  + \log N(w_i; 0, \tau_1^2 ) \right) + (1 - p_i)\left( \log \frac{1- \pi}{1 - p_i}  + \log N(w_i; 0, \tau_0^2 ) \right). 
\end{align*}
Thus, 
\begin{align}
    \text{(II)} & = p_i \left( \log \frac{\pi}{p_i}  +\mathbb{E}_{q_{\theta} (\mathbf{W})} \log \frac{N(w_i; 0, \tau_1^2)}{N(w; m_i, \sigma_i^2)} \right)  \nonumber \\  
    & \quad + (1 - p_i)\left( \log \frac{1- \pi}{1 - p_i}  +\mathbb{E}_{q_{\theta}(\mathbf{W})}\log \frac{N(w_i; 0, \tau_0^2)}{N(w; m_i, \sigma_i^2)}\right)   \label{eq:mid}.
\end{align}
To simplify this, recall the following fact about Gaussian Distribution
\begin{align*}
    \frac{N(X; a_1, b_1^2)}{N(X; a_2, b_2^2)}&= \frac{b_2}{b_1} \exp \Bigg\{-\frac{1}{2}  \left [ \left(\frac{1}{b_1^2} - \frac{1}{b_2^2}\right)  \left(X - \frac{\frac{a_1}{b_1^2} - \frac{a_2}{b_2^2}}{\frac{1}{b_1^2} - \frac{1}{b_2^2}}  \right)^2  + \frac{(a_1 - a_2)^2}{b_1^2 - b_2 ^2}  \right ]  \Bigg\}. 
\end{align*}
Using this, we can check 
\begin{align}
    \mathbb{E}_{q_{\theta}(\mathbf{W})} \log \frac{N(w_i; 0, \tau_1^2)}{N(w; m_i, \sigma_i^2)} 
    &= \log \frac{\sigma_i}{\tau_1} - \frac{m_i^2 + \sigma_i^2}{2\tau_1^2} + constant  \label{eq:mid2} \\
    \mathbb{E}_{q_{\theta}(\mathbf{W})} \log \frac{N(w_i; 0, \tau_0^2)}{N(w; m_i, \sigma_i^2)}
   & = \log \frac{\sigma_i}{\tau_0} - \frac{m_i^2 + \sigma_i^2}{2\tau_0^2} + constant \label{eq:mid3}.
\end{align}
Using \eqref{eq:mid}, \eqref{eq:mid2}, and \eqref{eq:mid3},  we get
\begin{align*}
  &\mathbb{E}_{q(\mathbf{W})} \mathbb{E}_{q(\mathbf{Z})} \log \frac{ \pi(\mathbf{W}, \mathbf{Z})}{q (\mathbf{W}) \cdot q(\mathbf{Z})} \\
     = & \sum_i \mathbb{E}_{q(\mathbf{W})} \left( \mathbb{E}_{q(\mathbf{Z})} \left(  \log  \pi(W_i, Z_i) - \log q(Z_i) \right)  - \log q(W_i) \right) \\
     = & \sum_i \left \{p_i \left( \log \frac{\pi}{p_i}  +\mathbb{E}_{q(\mathbf{W})} \log \frac{N(w_i; 0, \tau_1^2)}{N(w; m_i, \sigma_i^2)} \right) + (1 - p_i)\left( \log \frac{1- \pi}{1 - p_i}  +\mathbb{E}_{q(\mathbf{W})}\log \frac{N(w_i; 0, \tau_0^2)}{N(w; m_i, \sigma_i^2)}\right)  \right \} \\
     = & \sum_i \left\{p_i  \left( \log \frac{\pi}{p_i}  +\log \frac{\sigma_i}{\tau_1} - \frac{m_i^2 + \sigma_i^2}{2\tau_1^2} \right) + (1 - p_i)\left( \log \frac{1- \pi}{1 - p_i}  +\log \frac{\sigma_i}{\tau_0} - \frac{m_i^2 + \sigma_i^2}{2\tau_0^2}\right)\right\} + constant \\
     = & -\sum_i \mathcal{R}(\theta_i, p_i) + constant,
\end{align*}
which completes the proof.

\section*{S2: Proof for Proposition \ref{lemma:p:function}}

The first term in $  \mathcal{J}(\theta, \textbf{p})$ is free of $\textbf{p}$. Thus, 
\begin{align*}
   \frac{\partial \mathcal{J} (\theta, \textbf{p})}{\partial p_i} = \frac{\partial \mathcal{R}(\theta, \textbf{p})}{\partial p_i} = \left(\frac{m_i^2 + \sigma_i^2}{2\tau_1^2} - \frac{m_i^2 + \sigma_i^2}{2\tau_0^2}\right) + \left(\log \frac{\tau_1}{\pi} - \log \frac{\tau_0}{1 - \pi} \right) + \log \frac{p_i}{1 - p_i}.
\end{align*}
So to optimize w.r.t $p_i$, we set: 
\begin{align*}
  &\frac{\partial \mathcal{J} (\theta, \textbf{p})}{\partial p_i} = 0 \\
\implies &\left(\frac{\tilde{m}_i^2 + \tilde{\sigma}_i^2}{2\tau_1^2} - \frac{\tilde{m}_i^2 + \tilde{\sigma}_i^2}{2\tau_0^2}\right) + \left(\log \frac{\tau_1}{\pi} - \log \frac{\tau_0}{1 - \pi} \right) + \log \frac{\tilde{p}_i}{1 - \tilde{p}_i} = 0 \\
\implies &\log \frac{\tilde{p}_i}{1 - \tilde{p}_i} = B_i - A_i\\
\implies& \tilde{p}_i= \frac{1}{1 + \exp \{ A_i - B_i \}}.
\end{align*}

\section*{S3: Derivation in Section \ref{sec:compare:obj}}
The quantity of interest in $\mathcal{J}_B$ \citep{blundell2015weight}, is
\begin{align*}
    f(W , m) = - \log \pi (W) + \log q (W| m).
\end{align*}
Its gradient with respect to $m$ can be found by using the reparametrization trick as outlined in Equation 3 in Section 3.2 of \cite{blundell2015weight}: 
\begin{align*}
    \nabla_m f &= \frac{\partial f(W,m)}{\partial W} \frac{\partial W}{\partial m} + \frac{\partial f(W,m)}{\partial m}\\ 
    &= - \frac{\partial }{\partial W}  \log \pi (W)  + \frac{\partial }{\partial W}  \log q (W| m) +  \frac{\partial}{\partial m} \log q (W| m) \\
    & = - \frac{\partial }{\partial W}  \log \pi (W), 
\end{align*}
where we use the facts that $W = m + \sigma \epsilon $ and therefore$ \frac{\partial W}{\partial m} = 1$, and the last equality is due to
$$
\frac{\partial }{\partial m}  \log q (W| m) = - \frac{(W - m)}{\sigma^2} = - \frac{\partial }{\partial W}  \log q (W| m).
$$
Hence
\begin{align*}
    \nabla_m f =  - \frac{\partial }{\partial W}  \log \pi (W) 
    = \Bigg(\frac{W}{\tau_1^2} \frac{\pi_1(W)}{\pi(W)} + \frac{W}{\tau_0^2} \frac{\pi_0(W)}{\pi(W)} \Bigg), \label{eq:grad_bl_1}
\end{align*}
where we have used the notation: $\pi(W) = \pi_1 (W) + \pi_0 (W)$ and
$$
    \pi_1(W) = \pi \cdot N (W | 0, \tau_1^2), \quad 
       \pi_0(W) = (1 - \pi) \cdot N (W | 0, \tau_0 ^2).
$$
Next, we check the gradient with respect to $\sigma^2$: 
\begin{align*}
    \nabla_{\sigma^2} f &= \frac{\partial f(W,m)}{\partial W} \frac{\partial W}{\partial \sigma^2} + \frac{\partial f(W,m)}{\partial \sigma^2}\\ 
    &= \left(- \frac{\partial }{\partial W}  \log \pi (W)  + \frac{\partial }{\partial W}  \log q (W) \right)  \frac{\partial W}{\partial \sigma^2}+  \frac{\partial}{\partial \sigma^2} \log q (W| \sigma^2). 
\end{align*}
Each term can be calculated and simplified to obtain:
\begin{align*}
    \nabla_{\sigma^2} f &= \left( \left(\frac{W}{\tau_1^2} \frac{\pi_1(W)}{\pi(W)} + \frac{W}{\tau_0^2} \frac{\pi_0(W)}{\pi(W)}\right) - \frac{W -m}{\sigma^2} \right) \frac{\epsilon}{2\sigma}
- \left( \frac{1}{2\sigma^2}
+ \frac{\epsilon^2}{2\sigma^2} \right) \\
    &= \left(W \Bigg(\frac{1}{\tau_1^2} \frac{\pi_1(W)}{\pi(W)} + \frac{1}{\tau_0^2} \frac{\pi_0(W)}{\pi(W)} - \frac{1}{\sigma^2} \Bigg) + \frac{m}{\sigma^2} \right) \frac{\epsilon}{2\sigma}
- \left(\frac{1}{2\sigma^2}
+ \frac{\epsilon^2}{2\sigma^2} \right)  \\
&= \frac{1}{2} \left(\frac{m}{\sigma}  \epsilon + \epsilon^2 \right )   \Bigg(\frac{1}{\tau_1^2} \frac{\pi_1(W)}{\pi(W)} + \frac{1}{\tau_0^2} \frac{\pi_0(W)}{\pi(W)} - \frac{1}{\sigma^2} \Bigg) +\frac{1}{2\sigma^2} \left(  (\epsilon^2 -1 ) + \frac{m}{\sigma} \epsilon \right).
\end{align*}
}

\newpage

\bibliographystyle{ba}
\bibliography{mybib}

\begin{acknowledgement}

\end{acknowledgement}

\end{document}